\documentclass{article}
\usepackage{iclr2027_conference,times}

\usepackage{amsmath,amsfonts,bm}

\def\Figref#1{Figure~\ref{#1}}

\def\Secref#1{Section~\ref{#1}}

\def\secrefs#1#2#3{sections \ref{#1}, \ref{#2} and \ref{#3}}
\def\eqref#1{equation~\ref{#1}}

\def\1{\bm{1}}

\def\vone{{\bm{1}}}
\def\vmu{{\bm{\mu}}}

\def\vb{{\bm{b}}}
\def\vc{{\bm{c}}}

\def\vl{{\bm{l}}}
\def\vm{{\bm{m}}}

\def\vo{{\bm{o}}}

\def\vr{{\bm{r}}}
\def\vs{{\bm{s}}}

\def\vu{{\bm{u}}}

\def\vx{{\bm{x}}}
\def\vy{{\bm{y}}}
\def\vz{{\bm{z}}}

\def\mA{{\bm{A}}}
\def\mB{{\bm{B}}}
\def\mC{{\bm{C}}}
\def\mD{{\bm{D}}}

\def\mG{{\bm{G}}}

\def\mI{{\bm{I}}}
\def\mJ{{\bm{J}}}

\def\mL{{\bm{L}}}
\def\mM{{\bm{M}}}

\def\mQ{{\bm{Q}}}
\def\mR{{\bm{R}}}
\def\mS{{\bm{S}}}

\def\mW{{\bm{W}}}

\def\mLambda{{\bm{\Lambda}}}
\def\mSigma{{\bm{\Sigma}}}

\DeclareMathAlphabet{\mathsfit}{\encodingdefault}{\sfdefault}{m}{sl}
\SetMathAlphabet{\mathsfit}{bold}{\encodingdefault}{\sfdefault}{bx}{n}

\newcommand{\E}{\mathbb{E}}

\newcommand{\R}{\mathbb{R}}

\usepackage[utf8]{inputenc} %
\usepackage[T1]{fontenc}    %
\usepackage[hidelinks]{hyperref}       %
\usepackage{url}            %
\usepackage{booktabs}       %
\usepackage{amsfonts}       %
\usepackage{nicefrac}       %
\usepackage[expansion=false]{microtype}      %
\usepackage{xcolor}         %
\usepackage{graphicx}
\usepackage{amsmath}
\usepackage{amsthm} 
\usepackage{color, soul}
\usepackage[inline]{enumitem} 
\usepackage{bm}

\newtheorem{theorem}{Theorem}
\newtheorem{corollary}[theorem]{Corollary}
\newtheorem{proposition}[theorem]{Proposition}

\theoremstyle{definition}
\newtheorem{assumption}{Assumption}
\newtheorem{definition}{Definition}

\theoremstyle{remark}
\newtheorem{remark}{Remark}

\newcommand{\vepsilon}{{\bm{\varepsilon}}}  %
\newcommand{\veta}{{\bm{\eta}}}             %
\newcommand{\vxi}{{\bm{\xi}}}               %
\newcommand{\vsigma}{{\bm{\sigma}}}         %
\newcommand{\mPi}{{\bm{\Pi}}}               %
\newcommand{\ybar}{\bar{\vy}}               %
\newcommand{\indep}{\perp\!\!\!\perp}       %
\newcommand{\ind}{\mathbb{I}}               %

\makeatletter
\renewcommand{\eqref}[1]{\textup{\tagform@{\ref{#1}}}}
\makeatother

\renewcommand{\Figref}[1]{Fig.~\ref{#1}}

\renewcommand{\Secref}[1]{Sec.~\ref{#1}}

\renewcommand{\secrefs}[3]{Secs.~\ref{#1}, \ref{#2}, and~\ref{#3}}
\newcommand{\Secsref}[2]{Secs.~\ref{#1}--\ref{#2}}
\newcommand{\Appref}[1]{App.~\ref{#1}}
\newcommand{\Appsref}[2]{Apps.~\ref{#1} and~\ref{#2}}
\newcommand{\Thmref}[1]{Theorem~\ref{#1}}

\newcommand{\Corref}[1]{Corollary~\ref{#1}}
\newcommand{\Propref}[1]{Proposition~\ref{#1}}
\newcommand{\Asmref}[1]{Assumption~\ref{#1}}
\newcommand{\Defref}[1]{Definition~\ref{#1}}
\newcommand{\Remref}[1]{Remark~\ref{#1}}

\newif\ifpreprint
\preprinttrue

\ifpreprint
  \iclrfinalcopy
\fi

\title{Identifying Neural Dynamics Using Nonlinear Interventional State Space Models}

\author{%
  Hanna M.~Tolle\\
  Department of Computing\\
  Imperial College London\\
  London, UK \\
  \texttt{hmt23@ic.ac.uk} \\
  \And
  Carles Balsells-Rodas\\
  Department of Computing\\
  Imperial College London\\
  London, UK \\
  \texttt{cb221@ic.ac.uk} \\
  \And
  Yixin Wang\\
  Department of Statistics\\
  University of Michigan\\
  Ann Arbor, USA \\
  \texttt{yixinw@umich.edu} \\
  \And
  Pedro A. M. Mediano\thanks{These authors jointly supervised this work.}\\
  Department of Computing\\
  Imperial College London\\
  London, UK \\
  \texttt{p.mediano@imperial.ac.uk} \\
  \And
  Amin Nejatbakhsh\footnotemark[1]\\
  Center for Neural Science\\
  New York University\\
  New York, USA \\
  \texttt{mn2986@nyu.edu} \\
}

\begin{document}
\maketitle
\ifpreprint
  \lhead{Preprint. Under review.} %
\fi
\begin{abstract}
State-space models (SSMs) recover low-dimensional latent dynamics from high-dimensional observations, but latents are generally not identifiable from observational data alone. Interventional state-space models (iSSMs) use perturbations to break observational symmetries and identify parameters up to permutation, scaling, and shift. However, prior work restricts identifiability to linear latent dynamics and piecewise-linear emissions, excluding many behaviors characteristic of neural circuits. We extend iSSMs to a substantially broader model class: we prove that (i) affine identifiability holds under general injective emission functions and Gaussian latent transitions, and (ii) with interventions, identifiability up to permutation, scaling, and shift is preserved under general nonlinear latent dynamics. We validate our theory on simulated FitzHugh-Nagumo dynamics, and then demonstrate two applications: adjudicating between competing models of decision-making; and predicting responses to unseen micro-stimulation in macaque and human neural recordings, with proof-of-concept generalization to a held-out stimulation site in the latter. Finally, simulations yield practical guidance for designing stimulation protocols that improve identifiability.

\end{abstract}

\section{Introduction}
In biology, and especially neuroscience, experimental interventions are central to answering causal questions. 
Neuroscientists routinely perturb circuits using pharmacology~\citep{bolz1989pharmacological}, electrophysiology~\citep{cohen2004electrical, histed2013insights}, photostimulation~\citep{fenno2011development, adesnik2021probing}, and more recently brain-machine interfaces~\citep{motiwala2026brain}.

Classical perturbation experiments establish causal involvement -- e.g., by silencing a brain region and measuring behavioral deficits. Many core questions in systems neuroscience, however, require mechanistic explanations that specify not only which variables are involved, but also how they interact to generate observed dynamics. Such explanations could resolve longstanding debates about circuit motifs of working memory~\citep{javadzadeh2024dynamic, daie2023feedforward, inagaki2019discrete}, e.g., or dynamical regimes underlying perceptual decision-making~\citep{galgali2023residual}. %

A standard approach to such mechanistic questions is to fit state-space models (SSMs) to neural recordings. SSMs posit a low-dimensional latent state with explicit dynamics, and an emission map linking latents to observed activity. Without identifiability, however, distinct mechanisms can produce identical observations, so a fitted SSM cannot generally be viewed as the true causal mechanism~\citep{qian2024partial, maheswaranathan2019universality, lipshutz2024disentangling, galgali2023residual}.

\begin{figure}[t]
    \centering
    \includegraphics[width=\linewidth]{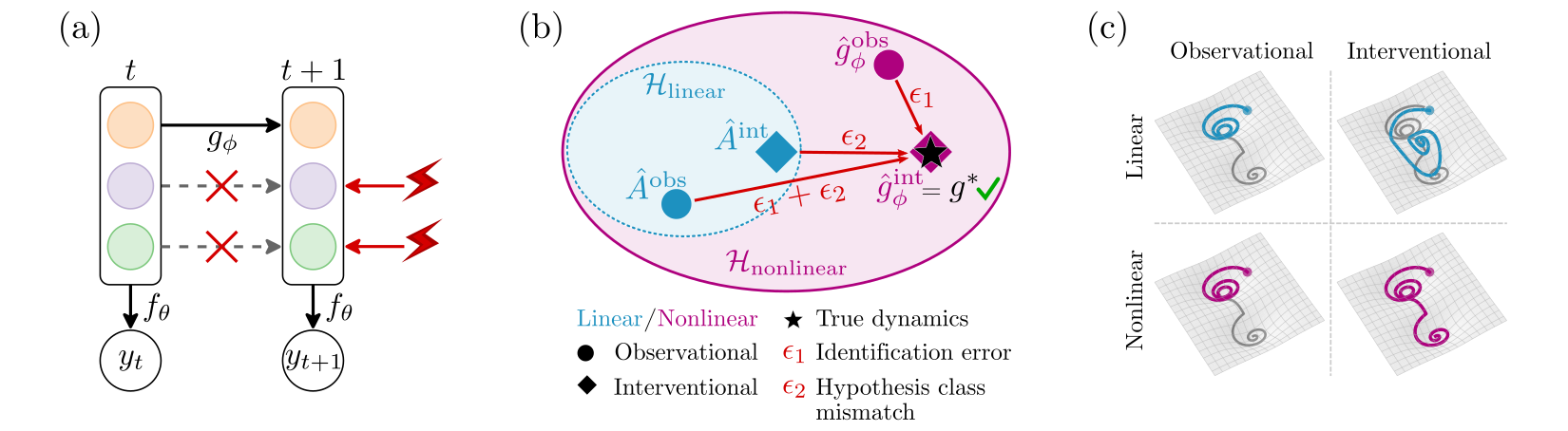}
    \caption{\textbf{Overview: ingredients for identifying latent dynamics in state-space models.}
    \textbf{(a)} iSSM leverages $\mathrm{do}$-interventions to identify latent dynamics up to permutation, coordinate-wise scaling, and shift.
    \textbf{(b)} Two independent failure modes prevent SSMs from identifying the true dynamics $g^*$: identification error $\epsilon_1$, resolved by interventions, and hypothesis class mismatch $\epsilon_2$, resolved by a nonlinear model class.
    \textbf{(c)} Latent trajectory recovery across model class (rows) and data regime (columns): only a nonlinear model trained on interventional data recovers the true trajectory (gray).
    }
    \label{fig:overview}
\end{figure}

Recent work by~\cite{nejat_wang_2025} addressed this gap with interventional state-space models (iSSMs), a framework for fitting SSMs to interventional data with theoretical guarantees of unique recovery up to benign symmetries. Although prior work has fitted SSMs to perturbation data, conventional SSMs remain subject to observational symmetries that permit spurious mechanisms. iSSM breaks those symmetries by fixing targeted latents to external inputs via $\mathrm{do}$-interventions (\Figref{fig:overview}a), severing causal dependencies from past latents and enabling identification up to permutation, coordinate-wise scaling, and shift (\citet{ahuja2023interventional} for a review of interventional inference).

While iSSM marked an important step, its guarantees only apply when the true system lies within the model's hypothesis class. Its identifiability results cover linear latent dynamics with piecewise-linear emission functions -- restrictive for neural circuits, which commonly exhibit nonlinear dynamics such as thresholding, bistability, and limit cycles. Specifically, we distinguish two inference failure modes (\Figref{fig:overview}b): (1) identification error $\epsilon_1$ from observational symmetries, resolved by interventional data; and (2) hypothesis class mismatch $\epsilon_2$ from model misspecification, requiring a sufficiently expressive nonlinear model class. \Figref{fig:overview}c illustrates both errors on a bistable-well system: without interventions, even a nonlinear model fails to recover transitions between attractors ($\epsilon_1$); with interventions, a linear model still misspecifies the true dynamics ($\epsilon_2$).

In this work, we generalize iSSM to nonlinear latent dynamics and general injective emission functions, addressing the principal open limitation of prior theory, and mitigating both inference failure modes -- $\epsilon_1$ through interventions and $\epsilon_2$ through a nonlinear hypothesis class. We validate the framework in simulations, demonstrate applications to macaque and human neural recordings under microstimulation, and distill practical guidance for experimental design to improve identifiability.

\section{Related Work}
This paper builds on the literature on identifiability in time series models. Broadly, this literature can be divided into theoretical guarantees obtained with or without interventional data.

\paragraph{Identifiability in time series models without interventions.} One line of work establishes identifiability purely from the statistical structure of latent processes. Early nonlinear ICA results exploited changing source distributions across time segments~\citep{hyvarinen2016unsupervised} or latent discrete Markov states~\citep{halva2020hidden,halva2021disentangling}, but treated latents as non-interacting; \citet{yao2022temporally} later allowed general nonlinear time-delayed coupling, identifying latents up to permutation and component-wise transformations when transition distributions vary sufficiently with the past. In a parallel line on switching dynamical systems, \citet{balsells_rodas_2024} identified coupled latents with regime-switching nonlinear Gaussian transitions up to an affine map under piecewise-linear emissions, and \citet{balsells2026identifiability} extended this to multi-lag dependencies and refined the map to permutation, scaling, and shift when noise variances differ across regimes.
Across these observational approaches, identifiability relies on non-interacting latents or transition noise that is non-Gaussian or state-/regime-dependent; for coupled dynamics with additive Gaussian noise of constant variance -- the standard SSM assumption -- temporal structure alone cannot resolve rotational symmetries~\citep{yao2022temporally,lippe2022citris}.

\paragraph{Identifiability in time series models via interventions.} A second line obtains identifiability from interventions or exogenous inputs acting on latents. CITRIS~\citep{lippe2022citris} and iCITRIS~\citep{lippe2022causal} identify latent causal variables with general nonlinear stochastic dynamics from intervened sequences, but only their intervention-dependent parts, up to component-wise transformations, and only when intervention targets are known. \citet{zhang2025identifiable} identify latents up to scaling under deterministic linear or control-affine dynamics in a known controllable canonical form with sparse input matrices. Interventional state-space models (iSSMs)~\citep{nejat_wang_2025} make the target map learnable and identify full latents up to permutation, scaling, and shift, yet only for linear latent dynamics and piecewise-linear emissions inherited from \citet{balsells_rodas_2024}.
The above approaches to interventional guarantees for latent dynamics therefore either require known intervention targets or restrict the dynamics to linear or control-affine forms and -- in the case of iSSM -- piecewise-linear emissions.

Building on this literature, we relax iSSM's piecewise-linear emission and linear-dynamics restrictions while retaining permutation-scale-shift identifiability under hard latent interventions.

\section{Methods}

\subsection{General Interventional State Space Models}

We consider a general class of interventional state-space models (iSSMs) that capture latent dynamical systems under interventions. Let $\vx_t \in \R^D$ denote latent variables and $\vy_t \in \R^N$ the observations at time $t$. The model is defined as:
\begin{align}
\vy_t &\sim P(\vy_t \mid f_{\theta}(\vx_t)), \label{eq:general_issm_y}\\
\vx_{t+1} &= \vm_t \odot g_{\phi}(\vx_t) + \mB \vu_t + \vepsilon_t, \label{eq:general_issm_x}
\end{align}

where $f_{\theta}$ is an injective emission function, $g_{\phi}$ is the latent transition function, $\vu_t \in \R^M$ denotes interventional inputs, and the process noise $\vepsilon_t$ has finite first moments and is independent of $(\vx_t,\vu_t)$. Specifically, we use Gaussian process noise $\vepsilon_t \sim \mathcal{N}(0, \mQ)$ with positive-definite covariance $\mQ$. The matrix $\mB \in \R^{D \times M}$ maps interventions to latent variables. The inputs $\vu_t$ are exogenous and observed, i.e., independent of the latent states and of the process noise. The binary mask $\vm_t \in \{0,1\}^D$ encodes which latent variables are not intervened on at time $t$, i.e., it is 1 where $(\mB\vu_t)_i=0$ and otherwise 0. This masking mechanism implements \emph{hard interventions} in latent space, where intervened variables are disconnected from their causal parents in the past.
While we use a linear input map $\mB\vu_t$ in all our experiments, we note that our theory extends to arbitrary deterministic input maps $h(\vu_t)$, with $m_{t,j}=\ind\{h_j(\vu_t)=0\}$, under a mild support condition on $h$ (\Appref{app:nonlinear_inputs}).

The above formulation reduces to the original iSSM when $g_{\phi}(\vx_t) = \mA \vx_t$ is linear and the emission function satisfies additional structural constraints.

\subsection{Theoretical Results}
\label{sec:theory}

We provide sufficient conditions for identifiability of the latent dynamics in the proposed nonlinear iSSM. The goal is to recover the latent transition and emission functions up to permutation, scaling, and shift. The main challenge is establishing this under nonlinear latent dynamics and general injective emissions.

Our approach follows the original iSSM proof and proceeds in three steps: (1) identify the distribution of noiseless latent representations from observed data; (2) identify latent variables and transition dynamics up to an affine transformation; and (3) use interventional structure to further restrict the equivalence class to permutation, scaling, and shift. Step (1) follows the observation-model argument of prior work. We extend Steps (2) and (3) to general injective emission functions and nonlinear latent dynamics.

We begin by identifying the distribution of the noiseless latent representations from the observed data. We denote the noiseless signal as $\ybar_t = f_{\theta}(\vx_t)$.
Under the following standard regularity condition, this distribution can be recovered from the observed data.

\begin{assumption}[Bounded completeness]
\label{as:bounded_completeness}
The observation kernel is bounded complete: for any finite signed measure $\mu$ on $\R^N$,
\begin{equation*}
    \int P(\vy_t \in E \mid \ybar_t = \vz) \, d\mu(\vz) = 0 \quad \text{for every measurable } E\subseteq\R^N \;\; \Rightarrow \;\; \mu = 0.
\end{equation*}
\end{assumption}
This condition is satisfied by many commonly used observation models, including Gaussian additive noise and Poisson rate observations.

\begin{proposition}[Recovery of noiseless observations \citep{nejat_wang_2025}]
\label{prop:denoising_step}
Under \Asmref{as:bounded_completeness}, the distribution of $\{\ybar_t\}_{t=1}^T$ is uniquely identifiable from the observed distribution of $\{\vy_t\}_{t=1}^T$.
\end{proposition}

\Propref{prop:denoising_step} reduces identifiability to recovering latent variables from noiseless observations. 
This is needed because \Thmref{thm:affine_identifiability} requires noiseless observations $\{\ybar_t\}$, but experimental neural data is usually noisy.

For Step~(2), affine identifiability, prior work establishes this under injective piecewise-linear emissions~\citep{nejat_wang_2025}. We show the same conclusion for general injective emissions under Gaussian latent transitions and mild regularity.

\begin{theorem}[Affine identifiability]
\label{thm:affine_identifiability}
Consider the distribution of $\{\ybar_t\}_{t=1}^T$, where $\ybar_t = f_{\theta}(\vx_t)$. Assume the following conditions hold:
\begin{enumerate}[label=(A\arabic*)]
    \item \label{as:regularity} \textbf{Observation model and regularity.}
    The emission function $f_{\theta}$ is injective and continuously differentiable, with continuously differentiable inverse on its image. The latent transition function $g_{\phi}$ is continuously differentiable.
    
    \item \label{as:gaussian_latents} \textbf{Gaussian latent transitions.}
    The latent dynamics are Gaussian and follow a first-order Markov process:
    \begin{equation*}
        \vx_t \mid \vx_{t-1} \sim \mathcal{N}(g_{\phi}(\vx_{t-1}), \mQ), \quad \mQ\succ0.
    \end{equation*}
    
    \item \label{as:non_deg_dyn} \textbf{Local non-degeneracy of dynamics.} 
    There exists $\vx^*\in\R^D$ such that $\mJ_{g_{\phi}}(\vx^*)$ is full rank.

\end{enumerate}
Let $(f_{\hat{\theta}}, g_{\hat\phi}, \hat \mQ)$ be another model parametrisation under assumptions \ref{as:regularity}--\ref{as:non_deg_dyn} that induces the same distribution on $\{\ybar_t\}_{t=1}^T$, where $\ybar_t = f_{\hat{\theta}}(\hat{\vx}_t)$.
Then there exists an invertible affine transformation $\psi(\vx) = \mM\vx + \vc$ such that $\hat{\vx}_t = \psi(\vx_t)$ for all $t$, where $\mM \in \R^{D \times D}$ is invertible and $\vc \in \R^D$.
\end{theorem}

The proof of \Thmref{thm:affine_identifiability} is in \Appref{app:step2}.

We now show that the intervention-based reduction from affine to block identifiability extends to the general nonlinear iSSM, even without assuming faithfulness as required by the original iSSM proof~\citep{nejat_wang_2025}.

\begin{theorem}[Block identifiability under interventions]
\label{thm:block_identifiability}
Assume that observationally equivalent parameterizations are related by a common invertible affine map $\hat\vx_t=\mM\vx_t+\vc$, as established, e.g., by \Thmref{thm:affine_identifiability}.
Suppose additionally that the following conditions hold:
\begin{enumerate}[label=(B\arabic*)]
    \item \label{as:non_deg_dyn2} \textbf{Regularity and local non-degeneracy of dynamics.}
    The transition functions $g_\phi$ and $g_{\hat\phi}$ are continuously differentiable, and there exists $\vx^*\in\R^D$ such that $\mJ_{g_\phi}(\vx^*)$ is invertible.

    \item \label{as:full_latent_support} \textbf{Full state support across regimes.}
    For each intervention regime, including the intervention-free regime, the distribution of the current latent state $\vx_t$ has support $\R^D$.

\end{enumerate}
Then the affine transformation $\psi(\vx) = \mM\vx + \vc$ is block-diagonal up to permutation, i.e., each estimated latent coordinate depends only on a subset of true latent coordinates corresponding to an intervention block.
\end{theorem}

\begin{assumption}[Unordered pairs condition] %
\label{as:unordered_pairs}
A set of interventions satisfies the unordered pairs condition on the latents if, for any unordered pair of latent coordinates $(i,j)$ with $i \neq j$, there exists an intervention environment in which exactly one of the two coordinates is intervened on.
\end{assumption}

\begin{corollary}[Permutation, scaling, and shift identifiability]
\label{cor:perm_scale_shift}
Under the conditions of \Thmref{thm:block_identifiability}, assume further that the unordered pairs condition holds. Then the affine transformation $\psi(\vx) = \mM\vx + \vc$ reduces to a permutation, scaling, and shift of the latent coordinates. That is, $\mM$ is a scaled permutation matrix, and the latent variables are identifiable up to permutation, scaling, and shift.
\end{corollary}

Permutation-scale-shift identifiability is desirable because it enables prediction of intervention combinations never seen in training. Additionally, \Asmref{as:unordered_pairs} implies that experimenters can partially control latent disentanglement by adding intervention channels, which increases the chances of latent pairs for which the unordered-pairs condition holds.

\subsection{Evaluation metrics}
\label{sec:metrics}

\paragraph{Reconstruction accuracy.}
The encoder infers latents from the full observed sequence and decodes them to observations. Reconstruction accuracy is the Pearson correlation between predicted and true observations on held-out trials. Prior work argues that reconstruction-based metrics poorly reflect capture of underlying dynamics~\citep{versteeg2025computation, lampinen2026representation, sedler2023expressive}; we therefore report two additional metrics.

\paragraph{Rollout accuracy.}
We take only the initial latent state $\vx_0$ from the encoder and roll out the learned dynamics open-loop without further observations. On neural recordings, the encoder sees only pre-stimulation observations when inferring $\vx_0$ (\Appref{app:real_eval}). Rollout accuracy is the Pearson correlation between predicted and true observations over the horizon on held-out trials.

\paragraph{Latent accuracy.}
When ground-truth latents are available, we apply the same open-loop rollout in latent space. Since iSSM identifies latents only up to permutation, coordinate-wise scaling, and shift (\Corref{cor:perm_scale_shift}), we align inferred latents to ground truth before evaluation. We use the Hungarian algorithm to find the optimal permutation matching columns of inferred and true $\mB$ matrices, then fit per-dimension affine transforms (scaling and shift) from true to inferred latents by ordinary least squares on test trajectories. Latent accuracy is the Pearson correlation between aligned predicted and true latent trajectories on held-out trials.

\section{Results}
\label{sec:results}

We first validate our theoretical results on simulated FitzHugh-Nagumo dynamics with known ground truth (\Secref{sec:fhn}). We then demonstrate the framework's practical value across increasingly realistic use cases: adjudicating between competing models of decision-making under idealized conditions (\Secref{sec:dmm}); and predicting responses to unseen micro-stimulation in macaque dlPFC and human intracranial recordings, with proof-of-concept generalization to a held-out stimulation site in the human data (\Secsref{sec:monkey}{sec:huang}). Finally, we return to controlled simulations to ask how future experiments can improve identifiability through the design of their stimulation protocols (\Secref{sec:stim}).

\subsection{Identifying FitzHugh-Nagumo dynamics}
\label{sec:fhn}

\begin{figure}[h]
  \centering
  \includegraphics[width=\linewidth]{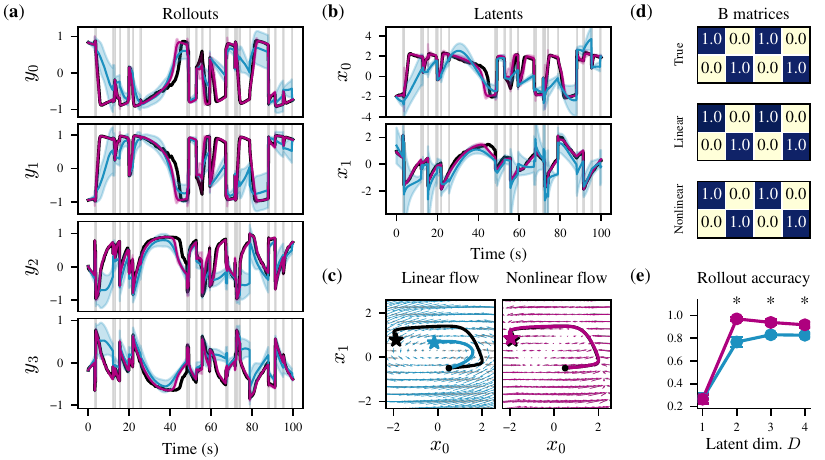}
  \caption{\textbf{Nonlinear iSSM identifies FitzHugh-Nagumo latent dynamics.}
  (a,b)~Open-loop rollout and aligned latent trajectories on a test trial: true (black), linear iSSM (blue), nonlinear iSSM (magenta); mean $\pm$ s.d.\ across 10 seeds. Gray lines mark interventions.
  (c)~Latent flow fields from the seed with highest Jacobian similarity to ground truth. Gray/black: true flow (arrows) and trajectory (line); blue/magenta: inferred linear/nonlinear flow and trajectories.
  (d)~Median inferred input map $\mB$ after permutation-scale alignment.
  (e)~Latent-dimension sweep over $D \in \{1,2,3,4\}$ shows that model selection picks out the true dimension ($D=2$).
  }
  \label{fig:fhn}
\end{figure}

The FitzHugh-Nagumo (FHN) system is a canonical nonlinear model of neuronal excitability with thresholding and stable limit-cycle dynamics~\citep{FitzHugh1961ImpulsesAP, Nagumo1962AnAP}. Because these dynamics lie outside the linear iSSM hypothesis class, FHN provides a suitable test of whether nonlinear iSSM can identify nonlinear latent dynamics from observations.

We simulated FHN dynamics
\begin{align}
    \frac{dv}{dt} &= v - \frac{v^3}{3} - w + I_\mathrm{ext}, &
    \tau \frac{dw}{dt} &= v + a - bw,
\end{align}
with $a = 0.7$, $b = 0.8$, $\tau = 12.5$, and $I_\mathrm{ext} = 0.5$ ($D=2$ latent dimensions). Latents were mapped to $N=4$ observations through a fixed random injective projection and pointwise $\tanh$ nonlinearity. Training used $K=200$ trials of length $T=50\,\mathrm{s}$ at $\Delta t=0.1\,\mathrm{s}$, with hard interventions arriving as a Poisson process at rate $\lambda=0.2\,\mathrm{s}^{-1}$ and amplitudes in $[-2,2]$. We trained 10 seeds per model class and evaluated on 100 held-out trajectories with novel interventions; full details are in \Appref{app:fhn_details}.

The nonlinear iSSM substantially outperforms the linear iSSM on both rollout and latent accuracy (\Figref{fig:fhn}a,b; both $p=0.0039$). Consistent with this performance gap, nonlinear models recover the characteristic relaxation-type limit cycle whereas linear models produce an elliptic spiral attractor (\Figref{fig:fhn}c); both recover the input map $\mB$ (\Figref{fig:fhn}d). Finally, a latent-dimension sweep over $D \in \{1,2,3,4\}$ selects the true dimension ($D=2$; \Figref{fig:fhn}e). We replicate this analysis on the Van der Pol oscillator in \Appref{app:vdp}. Overall, these results support \Corref{cor:perm_scale_shift}: nonlinear iSSM identifies the latent dynamics up to permutation, scaling, and shift, while linear iSSM cannot because the true system lies outside its hypothesis class.

\subsection{Identifying latent dynamics of decision-making models}
\label{sec:dmm}

We next consider a scientific use case under idealized conditions: adjudicating between three competing models of decision making. These posit (i) input-driven stable decay toward a point attractor, (ii) perfect integration along a line attractor, or (iii) autonomous nonlinear integration in a bistable well. Despite their distinct recurrent mechanisms, these systems can generate observationally indistinguishable trajectories~\citep[Fig.~1]{galgali2023residual}.

We simulated the three systems ($D=2$),
\begin{align}
    \dot{\vx} &= \mathrm{diag}(a_1,a_2)\,\vx && \text{(point attractor)}, \\
    \dot{\vx} &= \mA \vx && \text{(line attractor)}, \\
    \dot{x}_1 &= a x_1^3 + b x_1 + c_\mathrm{oh}, &
    \dot{x}_2 &= c x_2 + c_\mathrm{oh} && \text{(bistable well)},
\end{align}
with $(a_1,a_2)=(-0.5,-1.0)$, $\mA = \mathrm{Rot}(\gamma) \mR \mLambda \mR^{-1}$ ($\lambda_1=-1.0$, $\gamma=45^\circ$), and $(a,b,c,c_\mathrm{oh})=(-0.6,2.0,-1.0,0.1)$.
Following the protocol in \Secref{sec:fhn}, latents were mapped to $N=10$ observations; training used $K=40$ trials of length $T=20\,\mathrm{s}$ at $\Delta t=0.1\,\mathrm{s}$, with hard interventions at rate $\lambda=0.75\,\mathrm{s}^{-1}$ and amplitudes in $[-3,3]$ on $M=2$ input channels.
We trained 10 seeds per model class and evaluated on 100 held-out trajectories; full details are in \Appref{app:dmm_details}.

\begin{figure}[h]
    \centering
    \includegraphics[width=\linewidth]{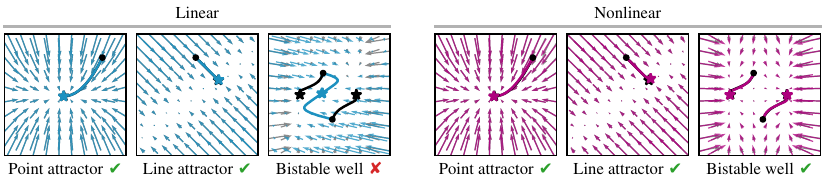}
    \caption{\textbf{Nonlinear iSSM identifies a broader class of decision-making dynamics.}
    Latent flow fields and trajectories for linear (left, blue) and nonlinear (right, magenta) iSSMs trained on interventional data from point-, line-, and bistable-well systems.
    Ground truth: gray arrows and black lines. Panels show best seed by Jacobian similarity; check marks indicate correct qualitative recovery.
    }
    \label{fig:dmm}
\end{figure}

As expected, the linear iSSM recovers the point- and line-attractor dynamics, which lie within its hypothesis class (\Figref{fig:dmm}, left). However for the bistable-well system, its learned flow collapses the two wells into a single fixed point and fails to reproduce trajectories on both sides of the separatrix. In contrast, the nonlinear iSSM recovers the flow fields and trajectories of all three systems (\Figref{fig:dmm}, right); and on the bistable system, it achieves significantly higher latent accuracy ($p=0.0044$).

Interventional data can disambiguate competing decision models, but only within the assumed dynamics class. Nonlinear iSSM extends that class to include bistable attractor dynamics that linear iSSM inherently cannot represent. We next show that this also matters for real neural recordings.

\subsection{Predicting macaque dlPFC responses to micro-stimulation}
\label{sec:monkey}

\begin{figure}[h]
  \centering
  \includegraphics[width=\linewidth]{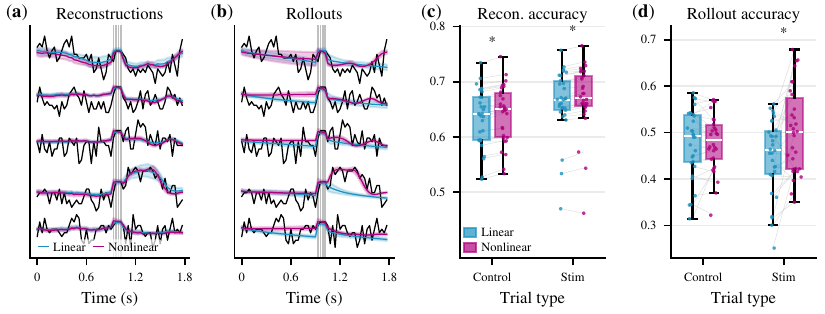}
  \caption{\textbf{Nonlinear iSSM outperforms linear iSSM on macaque prefrontal cortex recordings.}
  (a,b)~Example reconstructions and rollouts: observations (black), linear iSSM (blue), nonlinear iSSM (magenta); mean $\pm$ s.d.\ across seeds. Gray lines mark interventions.
  (c,d)~Reconstruction and rollout accuracy on held-out control and stimulation trials.
  Asterisks: $p < 0.05$.
  }
  \label{fig:kiani}
\end{figure}

We next show how iSSM can be used to predict neural responses to unseen micro-stimulation of known intervention sites. We applied linear and nonlinear iSSMs to electrophysiological recordings from macaque dlPFC~\citep{Nejatbakhsh2023PredictingTE}, pooling data from 14 sessions of the same monkey comprising 686 control and 685 interventional trials across $N = 96$ electrodes and $M = 12$ stimulation sites ($D = 12$; \Figref{fig:kiani}a). Training details are in \Appref{app:kiani_details}.

We separate evaluation by trial type -- control versus stimulation -- to distinguish modelling of spontaneous dynamics from generalisation to unseen interventions. The nonlinear iSSM significantly outperforms the linear iSSM on reconstruction accuracy on both control and stimulation trials (\Figref{fig:kiani}c). On rollout accuracy, the largest gap appears under stimulation (\Figref{fig:kiani}d).

These results suggest that dlPFC dynamics lie outside the linear iSSM hypothesis class.
The rollout gap under stimulation, but not control, implies this matters most for predicting evoked responses.

\subsection{Predicting human iEEG responses to single-pulse stimulation}
\label{sec:huang}

\begin{figure}[h]
  \centering
  \includegraphics[width=\linewidth]{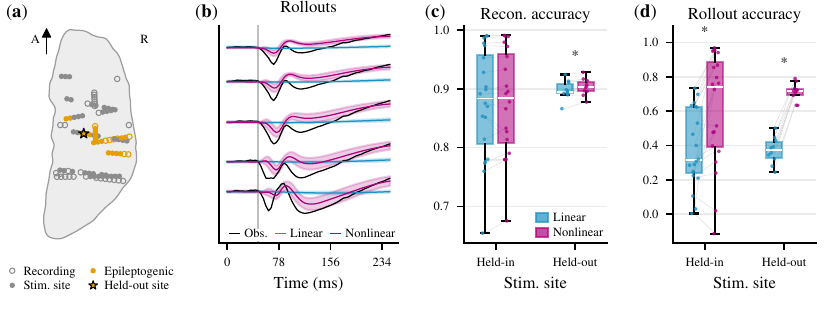}
  \caption{\textbf{Nonlinear iSSM predicts human iEEG stimulation responses more accurately, including at a held-out site.}
  (a)~Axial view: recording contacts, stimulation sites, annotated epileptogenic contacts, and the held-out site.
  (b)~Rollouts for the held-out site trial: observations (black), linear iSSM (blue), nonlinear iSSM (magenta); mean $\pm$ s.d.\ across seeds. Grey line: stimulation pulse.
  (c,d)~Reconstruction and rollout accuracy on held-in and held-out site trials.
  Asterisks: $p < 0.05$.}
  \label{fig:huang}
\end{figure}

We apply linear and nonlinear iSSMs to intracranial EEG from one epilepsy patient recorded during single-pulse electrical stimulation~\citep{huang2023electrical} ($N = 99$ channels, $M = 50$ sites, $K=423$ trials, $D = 25$; \Figref{fig:huang}a), extending the macaque analysis to a second species and recording modality. Test trials fall into two groups: 20 from sites seen during training; and 10 at a site, clinically labelled irritative zone, fully withheld from training. To enable predicting stimulation response at this site, never seen during training, we map each site's input column from electrode coordinates via a site-embedding network (\Appref{app:huang}).

On held-out trials from trained stimulation sites, both models reconstruct equally well, but the nonlinear iSSM predicts rollouts significantly better ($p = 0.0004$; \Figref{fig:huang}b--d). At the held-out site, the nonlinear iSSM reaches rollout accuracy $0.71 \pm 0.02$ versus $0.37 \pm 0.02$ for the linear iSSM ($p = 0.002$; mean $\pm$ s.e.m. across held-out trials), with a small reconstruction advantage. 

Note that our theory constrains the input map only on observed stimulation sites (\Remref{rem:input_map_support}); thus, the held-out-site evaluation tests coordinate-based interpolation and is included here only as a proof of concept. What makes this empirical generalization test possible is that the held-out site lies 3.5~mm from the nearest held-in site, and recorded responses provide ground truth for validation.

\subsection{Designing interventions and evaluating latent recovery}
\label{sec:stim}

Our theory offers practical guidance on how future stimulation experiments can be designed to maximize identifiability: the theorems are population statements requiring the observation law to be matched over the latent support, so identification holds only where training trajectories visit. Thus, latent accuracy should improve with intervention number and diversity. We test this on the bistable well system (details in \Appref{app:stim_details}), and replicate on two other systems (\Appref{app:stim_point_line}).

\begin{figure}[h]
    \centering
    \includegraphics[width=\linewidth]{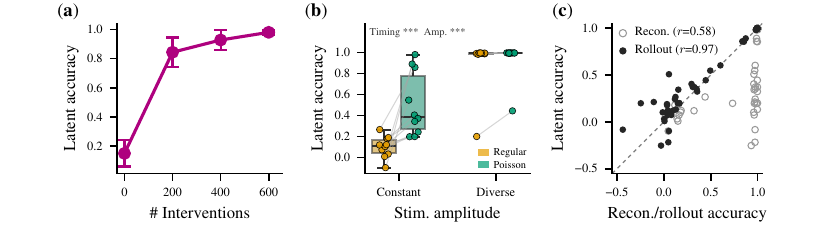}
    \caption{\textbf{Numerous and diverse interventions improve latent recovery.}
    (a)~Latent accuracy improves with more training interventions.
    (b)~Diverse amplitudes and irregular timing yield higher latent accuracy; boxes: 10 seeds, ***$p<0.001$ main effect.
    (c)~Latent accuracy versus rollout (black) and reconstruction (gray) accuracy; gray dashed line: identity.
    }
    \label{fig:interventions}
\end{figure}

First, we sweep the Poisson stimulation rate over $\lambda \in \{0, 0.25, 0.5, 0.75\}\,\mathrm{s}^{-1}$, generating $0$ to ${\sim}600$ interventions for training (\Figref{fig:interventions}a). Training nonlinear iSSM on each condition shows latent accuracy increases monotonically with the number of interventions from near chance to near ceiling.
Next, we compare stimulation protocols, crossing (constant or diverse) amplitude with (regular or Poisson) timing at a matched total number of pulses. Amplitude diversity has the larger main effect on latent accuracy (Cohen's $d=1.86$, $p=5.7\times 10^{-6}$), although irregular timing also helps ($d=0.71$, $p=1.4\times 10^{-5}$; \Figref{fig:interventions}b). The fixed-amplitude, regular-timing protocol performs worst.

As ground-truth latents are typically unavailable in real data, we ask which observable metric best reflects latent recovery. Pooling all seeds and conditions from \Figref{fig:interventions}a--b, rollout accuracy tracks latent accuracy closely (points scatter around the identity line), whereas the more commonly used reconstruction accuracy stays near ceiling even when latent recovery fails (\Figref{fig:interventions}c).

\section{Discussion}
We generalize iSSM identifiability guarantees to injective emissions and nonlinear latent dynamics. We demonstrate two applications -- adjudicating between decision-making models, and predicting stimulus response in real neural recordings -- which speak to central neuroscience objectives.

\paragraph{Implications for neuroscience.}
A core objective in systems neuroscience is to identify dynamic regimes of cognition: e.g., prior work used dynamical inference on neural recordings to characterize evidence accumulation for decision-making~\citep{latimer2015single, luo2025transitions}. Yet, conclusions remain contested~\citep{Zoltowski2018DiscreteSA, Steinemann2024DirectOO}; in part because equally good fits can imply different dynamics, which led \citet{Genkin2020MovingBG} to propose an empirical model selection approach that prioritizes consistency of inferred dynamics. Our framework offers a complementary route that combines formal identifiability guarantees with considerable expressiveness.

A second objective, with direct clinical relevance, is predicting neural response to stimulation -- e.g., to localize seizure onset zones in epilepsy or to optimize deep brain stimulation targets. Whole-brain biophysical models pursue this via virtual stimulations~\citep{Hashemi2023AmortizedBI, Deco2014GreatEU}. However, they are strongly hypothesis-driven, commonly fitting only a few parameters to data~\citep{Deco2008TheDB, Deco2018WholeBrainMN}, and current state-of-the-art models achieve comparatively poor fits~\citep{Herzog2024NeuralMM}. iSSM offers a data-driven alternative, and we show that it predicts stimulus response in both macaque and human intracranial recordings, and enables response prediction at unseen stimulation sites. Notably, an exploratory analysis suggests that virtual stimulations, including at several untrained sites, may discriminate healthy from epileptogenic tissue (\Appref{app:huang_virtual_stim}); though this extends beyond our identifiability guarantees (see Limitations).

\paragraph{Limitations.}
\Thmref{thm:block_identifiability} assumes hard interventions. If microstimulation is not perfectly hard, latent and rollout accuracy may degrade due to misspecification. Nevertheless, toy-model simulations show partial robustness as interventions soften (\Appref{app:intervention_hardness}; \Figref{fig:baselines}); and our theory requires only stochastically hard interventions rather than a deterministic clamp to a known value. Established techniques such as dynamic clamps~\citep{Prinz2004TheDC} or optogenetic silencing~\citep{Wiegert2017SilencingNT} plausibly operate within this regime. Mapping experimental techniques onto hard/soft regimes is an important avenue for future work.

For stimulation-response prediction, identifiability holds only over the observed support of latent inputs (\Remref{rem:input_map_support}). Predicting response at unseen sites is therefore extrapolation that our theory does not guarantee. Additionally, we analysed data from only one patient; and the 99 shaft contacts are unlikely to sufficiently sample the full hemisphere. Reliable prediction at distant sites likely requires denser coverage and validation across multiple patients.

Stimulation artefacts, which corrupt signal at stimulated channels, pose a further practical challenge. In the human intracranial recordings, we mitigate this by interpolating and masking stimulated recording sites (\Appref{app:huang_data}), and we extend our identifiability theory to cover fitting under such observation masking (\Appref{app:obs_masking}).

\paragraph{Progressing towards improved identifiability in real data.}
Accurate identification of something as complex as neural dynamics needs adequate data. Typical interventional datasets, however, include only a few channels, fixed amplitudes, regularly timed pulses, and limited repetitions. Our toy-model simulations clearly show that more diverse and numerous stimulations substantially improve latent recovery. We hope that these results motivate richer interventional protocols to help advance towards identifying causal mechanisms of neural dynamics.

\section*{AI Use Statement}

In this work, we used generative AI tools (Claude Code and ChatGPT) to assist in the writing of proofs (with presentation and proof checking; see below), provide feedback on research methodology and experiments, interpret results, and via coding assistance: implement methods, generate synthetic data, clean and reformat datasets. We have not used generative AI tools to formulate mathematical claims, provide critical ingredients for proving mathematical claims, help develop theoretical models or conceptual frameworks, or propose or refine hypotheses, and translation and qualitative and thematic data analysis are not applicable to this work. Additionally, we used generative AI tools to create or edit software code, summarize existing literature, draft parts of the paper (such as this section) and edit it for readability. We have reviewed all AI-assisted work. All theoretical results and proofs were developed by the authors; AI was used to discuss proofs in related work, check our proofs for errors, and improve their clarity in the writeup. Experiments and methods were designed by the authors; AI gave feedback and assisted with coding, including implementing the preprocessing pipeline for the human iEEG data according to the pipeline published by the original authors~\citep{huang2023electrical}. All AI-assisted code was reviewed and tested by the authors. Results were interpreted by the authors, with AI used for discussion. AI drafted the ethics, reproducibility and AI use statements; the main text was written by the authors, with AI used for polishing, streamlining and typo elimination. We take responsibility for the final content of this work, including text, claims or artifacts produced with the aid of generative AI.

\section*{Ethics Statement}

This work uses only previously published neural recordings; no new data were collected. The human iEEG data are the publicly available, de-identified OpenNeuro dataset \texttt{ds004457}~\citep{huang2023electrical}, recorded under Mayo Clinic Institutional Review Board approval with written informed consent from each patient, as reported by the original authors. The macaque recordings were previously described by~\citet{Nejatbakhsh2023PredictingTE}. The authors declare no conflicts of interest.

\section*{Reproducibility Statement}

Code for all experiments is available at \url{https://anonymous.4open.science/r/issm-code-E235/}. 
Assumptions for our identifiability results are stated in \Secref{sec:theory}, with complete proofs in \Appref{app:identifiability_proofs}. For each experiment, the appendix details data generation or preprocessing, model architectures, training procedures and computational resources (\Appref{app:fhn_details}--\ref{app:stim_details}); statistical tests are described in \Appref{app:stats}. The human iEEG data are publicly available on OpenNeuro (\texttt{ds004457}).

\ificlrfinal
\section*{Author Contributions}

Conceptualization, H.M.T.\ and A.N.; 
Data curation, A.N.; 
Formal analysis, H.M.T.; 
Investigation, H.M.T.; 
Methodology, H.M.T.\ and C.B.-R. and Y.W.;
Software, H.M.T.\ and A.N.; 
Supervision, P.A.M.M.\ and A.N.; 
Validation, C.B.-R.\ and Y.W.\ and P.A.M.M.; 
Visualization, H.M.T.\ and C.B.-R.; 
Writing -- original draft, H.M.T.\ and A.N.; 
Writing -- review \& editing, H.M.T.\ and C.B.-R.\ and Y.W.\ and P.A.M.M.\ and A.N.

\section*{Acknowledgments}

H.M.T.\ is supported by the Doctoral Teaching Scholarship of the Department of Computing, Imperial College London. Y.W is supported in part by funding from the Office of Naval Research under grant N00014-23-1-2590, the Department of Energy under grant DE-SC0026902, the National Science Foundation under grant DMS-2310831, DMS-2428059, IIS-2435696, IIS-2440954, Two Sigma Investments LP, Science of Trustworthy AI program at Schmidt Sciences, and LG Management Development Institute AI Research.

\fi

{\small
\bibliographystyle{iclr2027_conference}
\bibliography{refs}
}

\appendix

\section{Identifiability Proofs}
\label{app:identifiability_proofs}

\subsection{Setup}
\label{app:proof_setup}

We consider the interventional state space model
\begin{align}
    \vy_t &\sim P(\vy_t \mid f_\theta(\vx_t)), \label{eq:setup_obs}\\
    \vx_{t+1} &= \vm_t \odot g_\phi(\vx_t) + \mB\vu_t + \vepsilon_t, \label{eq:setup_latent}
\end{align}
where $\vx_t \in \R^D$, $\vy_t \in \R^N$, 
$\vm_t \in \{0,1\}^D$ with $m_{t,j} := \ind\{(\mB\vu_t)_j = 0\}$ is the intervention mask, 
and $\odot$ denotes elementwise multiplication. The inputs $\vu_t$ are exogenous and observed, i.e., independent of the latent states and of the process noise.
The process noise $\vepsilon_t$ has finite first moments and is jointly independent of the current latent state $\vx_t$ and intervention $\vu_t$, i.e., $\vepsilon_t \indep (\vx_t, \vu_t)$.

Throughout, we compare two parameterizations from the same iSSM model class that induce identical observation distributions within each considered intervention regime, including the intervention-free regime $\vu=0$. We denote the second parameterization by hats:
\begin{align}
    \vy_t &\sim P(\vy_t \mid f_{\hat\theta}(\hat \vx_t)), \\
    \hat \vx_{t+1} &= \hat \vm_t \odot g_{\hat\phi}(\hat \vx_t) + \hat \mB \vu_t + \hat \vepsilon_t.
\end{align}
Both parameterizations share the same observation kernel $P(\vy_t \mid \cdot)$; that is, we compare candidate latent dynamics and decoders under a common measurement map, which is given by the recording process. Also note that observations are conditionally independent across time given the noiseless signals.

\paragraph{Notation.}
We use $P$ for probability laws and conditional distributions, and $p$ for densities where they exist. No density is assumed for the observations or noiseless signals in Step~1. Events $E$ and $E_t$ are measurable subsets of the observation space.

\subsection{Step 1: Identification of the noiseless representation}
\label{app:step1}

We follow the observation-model argument of \citet{nejat_wang_2025}, allowing noiseless distributions without densities and establishing identification of the noiseless transitions.

Define the noiseless signal $\ybar_t:=f_\theta(\vx_t)$. Write
\begin{equation}
K_t(E_t\mid \vz_t) := P(\vy_t\in E_t\mid\ybar_t=\vz_t)
\end{equation}
for the observation kernel, which is shared by both parameterizations. Here $\vz_t\in\R^N$ is a possible value of the noiseless signal $\ybar_t$. We use $E_t$ for measurable events in the observation space and $F_t$ for measurable events in the noiseless-signal space.

\paragraph{Noiseless marginals.}
For every measurable observation event $E$,
\begin{equation}
P(\vy_t\in E)=\int K_t(E\mid \vz)\,P_{\ybar_t}(d\vz).
\end{equation}
If two candidate parameterizations induce the same observed distribution $P_{\vy_t}=\hat{P}_{\vy_t}$, we have
\begin{equation}
\int K_t(E\mid \vz)\,\bigl(P_{\ybar_t}-\hat P_{\ybar_t}\bigr)(d\vz)=0
\qquad\text{for every }E.
\end{equation}
By \Asmref{as:bounded_completeness}, $P_{\ybar_t}=\hat P_{\ybar_t}$, so the noiseless marginal is identified.

\paragraph{Noiseless transitions.}
We first identify the joint law of two consecutive noiseless signals; their conditional distribution then follows. For convenience, we label the two time points as $1,2$, but they can be any two adjacent times.

Conditional independence of the observations given the noiseless signals yields
\begin{equation}
P(\vy_1\in E_1,\vy_2\in E_2)
=
\int K_1(E_1\mid \vz_1)K_2(E_2\mid \vz_2)\,
P_{\ybar_1,\ybar_2}(d\vz_1,d\vz_2).
\end{equation}
Let
\begin{equation}
\mu := P_{\ybar_1,\ybar_2}-\hat P_{\ybar_1,\ybar_2}
\end{equation}
be the finite signed measure given by the difference of the two candidate noiseless joint distributions. Since the candidates induce the same observed joint law, subtracting their expressions above gives
\begin{equation}
\int K_1(E_1\mid \vz_1)K_2(E_2\mid \vz_2)\,\mu(d\vz_1,d\vz_2)=0
\qquad\text{for every }E_1,E_2.
\label{eq:alt_zero_integral}
\end{equation}
We want to show that this implies $\mu=0$; and we will do this by applying \Asmref{as:bounded_completeness} once to each coordinate.

First, fix an observation event $E_2$ and define a signed measure on the first noiseless coordinate by
\begin{equation}
\nu_{E_2}(F_1)
:=
\int_{F_1\times\R^N}
K_2(E_2\mid \vz_2)\,\mu(d\vz_1,d\vz_2).
\end{equation}
This measure is finite because $\mu$ is finite and $0\leq K_2(E_2\mid \vz_2)\leq 1$. Substituting into Eq.~\eqref{eq:alt_zero_integral} gives
\begin{equation}
\int K_1(E_1\mid \vz_1)\,\nu_{E_2}(d\vz_1)=0
\qquad\text{for every }E_1.
\end{equation}
By \Asmref{as:bounded_completeness}, $\nu_{E_2}=0$. Since $E_2$ was arbitrary, this gives
\begin{equation}
\int_{F_1\times\R^N}
K_2(E_2\mid \vz_2)\,\mu(d\vz_1,d\vz_2)=0
\qquad\text{for every }F_1,E_2.
\label{eq:alt_sliced_integral}
\end{equation}

Next, we fix a noiseless event $F_1$ and define a finite signed measure on the second coordinate by
\begin{equation}
\rho_{F_1}(F_2):=\mu(F_1\times F_2).
\end{equation}
Eq.~\eqref{eq:alt_sliced_integral} then becomes
\begin{equation}
\int K_2(E_2\mid \vz_2)\,\rho_{F_1}(d\vz_2)=0
\qquad\text{for every }E_2.
\end{equation}
A second application of \Asmref{as:bounded_completeness} gives $\rho_{F_1}=0$. Since $F_1$ was arbitrary,
\begin{equation}
\mu(F_1\times F_2)=0
\qquad\text{for every }F_1,F_2.
\end{equation}
Thus the candidate joint distributions agree on all measurable rectangles, which determine a probability law on $\R^N\times\R^N$. Hence $\mu=0$, and the noiseless joint distribution is identified. Uniqueness of conditional distributions then identifies $P(\ybar_2\mid\ybar_1)$ up to a set of conditioning values of probability zero under the common marginal $P_{\ybar_1}$. Because the two time points were arbitrary, this identifies $P(\ybar_t\mid\ybar_{t-1})$ for every adjacent pair.

\paragraph{Noiseless trajectories.}
The same argument applies successively to every coordinate of a finite trajectory. Let
\begin{equation}
\mu_T := P_{\ybar_{1:T}}-\hat P_{\ybar_{1:T}}
\end{equation}
be the difference of the candidate noiseless trajectory laws. Equality of the full observed trajectory laws and conditional independence of the observations give
\begin{equation}
\int \prod_{t=1}^{T}K_t(E_t\mid \vz_t)\,\mu_T(d\vz_{1:T})=0
\qquad\text{for every }E_1,\dots,E_T.
\end{equation}
Fixing $E_2,\dots,E_T$, define a finite signed measure on the first coordinate by integrating the remaining kernel factors against $\mu_T$, as in the definition of $\nu_{E_2}$ above. Completeness of $K_1$ implies that this measure is zero on every noiseless event $F_1$. Next, fix $F_1$ and $E_3,\dots,E_T$ and apply completeness of $K_2$ to the corresponding signed measure on the second coordinate. Continuing in this way through all $T$ coordinates gives
\begin{equation}
\mu_T(F_1\times\cdots\times F_T)=0
\qquad\text{for every }F_1,\dots,F_T.
\end{equation}
The candidate distributions therefore agree on all measurable rectangles and hence on the entire noiseless trajectory distribution. This proves \Propref{prop:denoising_step}.

\subsection{Step 2: Affine identifiability with smooth injective decoders}
\label{app:step2}

We now provide an alternative route to affine identifiability that replaces the piecewise linearity 
assumption of \citet{nejat_wang_2025} by smoothness, injectivity, and Gaussian latent transitions.

\paragraph{Assumptions.}
We adopt the assumptions of \Thmref{thm:affine_identifiability}. In particular:
\begin{enumerate}[label=(A\arabic*)]
    \item The decoders are injective and continuously differentiable, with continuously differentiable inverses, and the latent transition function $g_\phi$ is continuously differentiable,
    \item the latent dynamics follow a Gaussian first-order Markov process with positive definite covariance, and
    \item there exists $\vx^*\in\R^D$ such that $\mJ_{g_{\phi}}(\vx^*)$ is full rank.
\end{enumerate}

\begin{proof}
Step~1 shows that observationally equivalent parameterizations induce the same noiseless transition distributions, almost everywhere under the common conditioning marginal, i.e.,
\begin{equation}
P(\ybar_t \mid \ybar_{t-1}) = \hat P(\ybar_t \mid \ybar_{t-1})
\end{equation}
almost surely.
Given injectivity of the decoder, we may pull back this equality to latent coordinates using $\vx_t = f_\theta^{-1}(\ybar_t)$ and $\hat \vx_t = f_{\hat \theta}^{-1}(\ybar_t)$:
\begin{equation}
    \mathrm{vol}\,\mJ_{f_\theta^{-1}}(\ybar_t)\, p(\vx_t \mid \vx_{t-1})
    =
    \mathrm{vol}\,\mJ_{f_{\hat\theta}^{-1}}(\ybar_t)\, \hat p(\hat \vx_t \mid \hat \vx_{t-1}).
\end{equation}

Invertibility of the decoders ensures that a mapping $\psi(\vx_t)=\hat \vx_t$ exists, where $\psi = f_{\hat \theta}^{-1} \circ f_\theta$. Using this definition and taking logarithms yields
\begin{equation}
    \log \mathrm{vol}\,\mJ_{f_\theta^{-1}}(\ybar_t) + \log p(\vx_t \mid \vx_{t-1})
    =
    \log \mathrm{vol}\,\mJ_{f_{\hat\theta}^{-1}}(\ybar_t) + \log \hat p(\psi(\vx_t) \mid \psi(\vx_{t-1})).
\end{equation}
Now apply the mixed derivative $\nabla_{\vx_t}\nabla_{\vx_{t-1}}$. The Jacobian-volume terms vanish because they depend on $\vx_t$ but not on $\vx_{t-1}$. Hence
\begin{equation}
    \nabla_{\vx_t}\nabla_{\vx_{t-1}} \log p(\vx_t \mid \vx_{t-1})
    =
    \nabla_{\vx_t}\nabla_{\vx_{t-1}} \log \hat p(\psi(\vx_t) \mid \psi(\vx_{t-1})).
    \label{eq:mixed_derivative_equality}
\end{equation}

For Gaussian latent transitions,
\begin{equation}
    \log p(\vx_t \mid \vx_{t-1})
    =
    -\frac{1}{2}(\vx_t - g_{\phi}(\vx_{t-1}))^\top \mQ^{-1} (\vx_t - g_{\phi}(\vx_{t-1})) + \mathrm{const},
\end{equation}
where the constant does not depend jointly on $(\vx_t,\vx_{t-1})$. Therefore,
\begin{equation}
    \nabla_{\vx_t}\nabla_{\vx_{t-1}} \log p(\vx_t \mid \vx_{t-1})
    = \mQ^{-1} \mJ_{g_{\phi}}(\vx_{t-1}).
\end{equation}
Analogously, for the hatted parameterization, the chain rule gives
\begin{equation}
    \nabla_{\vx_t}\nabla_{\vx_{t-1}} \log \hat p(\psi(\vx_t)\mid \psi(\vx_{t-1}))
    =
    \mJ_\psi(\vx_t)^\top \hat \mQ^{-1} \mJ_{g_{\hat\phi}}(\psi(\vx_{t-1})) \mJ_\psi(\vx_{t-1}).
\end{equation}
Substituting into \eqref{eq:mixed_derivative_equality} yields
\begin{equation}
    \mQ^{-1} \mJ_{g_{\phi}}(\vx_{t-1})
    =
    \mJ_\psi(\vx_t)^\top \hat \mQ^{-1} \mJ_{g_{\hat\phi}}(\psi(\vx_{t-1})) \mJ_\psi(\vx_{t-1}).
    \label{eq:key_jacobian_relation}
\end{equation}

By \Asmref{as:non_deg_dyn}, $\mJ_{g_{\phi}}(\vx^*)$ is full rank at some $\vx^*\in\R^D$. Since $g_{\phi}$ is continuously differentiable by \Asmref{as:regularity}, its Jacobian is continuous. Hence there exists a nonempty open neighbourhood $\mathcal{B}_{\vx^*}$ of $\vx^*$ on which $\mJ_{g_{\phi}}$ is invertible. Since all matrices in \eqref{eq:key_jacobian_relation} are of size $D \times D$, this implies that each factor on the right-hand side must also be full rank on that set. In particular, $\mJ_\psi(\vx_t)$, $\hat \mQ$, $\mJ_{g_{\hat\phi}}(\psi(\vx_{t-1}))$, and $\mJ_\psi(\vx_{t-1})$ are invertible for any $\vx_{t-1}\in\mathcal{B}_{\vx^*}$, and we may therefore rearrange the expression as
\begin{equation}
    \mJ_\psi(\vx_t)^\top =
     \mQ^{-1} \mJ_{g_{\phi}}(\vx_{t-1})
    \mJ_\psi(\vx_{t-1})^{-1}
    \mJ_{g_{\hat\phi}}(\psi(\vx_{t-1}))^{-1}
    \hat \mQ.
    \label{eq:jpsi_constant}
\end{equation}
By \Asmref{as:gaussian_latents} and for any $\vx_{t-1}\in\mathcal{B}_{\vx^*}$, the conditional distribution of $\vx_t|\vx_{t-1}$ has full support for $t>2$. Therefore, the above equation holds for any $\vx_t\in\R^D$. Since the right-hand side depends on $\vx_{t-1}$ but not on $\vx_t$, this implies  $\mJ_\psi(\vx_t)$ is constant with respect to $\vx_t$, and therefore, $\psi$ is affine:%
\begin{equation}
    \hat \vx_t =\psi(\vx_t) = \mM\vx_t + \vc,
\end{equation}
which proves affine identifiability.
Note that the decoders, and consequently $\psi=f_{\hat\theta}^{-1}\circ f_\theta$, do not depend on the intervention. The above affine map is therefore shared across regimes with and without interventions.
\end{proof}

\subsection{Step 3: Refinement under interventions}
\label{app:step3}

We now show how hard interventions refine affine identifiability (\Thmref{thm:affine_identifiability}) to block-wise permutation-scaling-shift identifiability. 

\paragraph{Notation.}
For an index set $\mathcal{J}\subseteq\{1,\dots,D\}$, $-\mathcal{J}$ denotes its complement. We write $\mM_{\mathcal{J}',\mathcal{J}}$ for the submatrix with rows indexed by $\mathcal{J}'$ and columns indexed by $\mathcal{J}$.

\paragraph{Assumptions.}
Step~3 assumes a common invertible affine correspondence $\hat\vx_t=\mM\vx_t+\vc$, regularity and local non-degeneracy of dynamics (\Asmref{as:non_deg_dyn2}), and full state support across intervention regimes (\Asmref{as:full_latent_support}).

\paragraph{Statement.}
For a fixed intervention input $\vu$, let $\mathcal{J}$ and $\mathcal{J}'$ be the target sets in the true and candidate parameterizations, respectively. We show that $|\mathcal{J}'|=|\mathcal{J}|$ and
\begin{equation}
    \mM_{\mathcal{J}',-\mathcal{J}}=0, \qquad
    \mM_{-\mathcal{J}',\mathcal{J}}=0.
\end{equation}
Thus, after aligning the coordinates, $\mM$ is block diagonal with invertible blocks, which means that the candidate coordinates in $\mathcal{J}'$ depend only on the true coordinates in $\mathcal{J}$. If $|\mathcal{J}|=1$, the intervened latent is identifiable up to permutation, scaling, and shift.

\begin{proof}
Let $\vb = \E[\vepsilon_t \mid \vx_t=\vx, \vu_t=\vu]$ and $\hat \vb = \E[\hat \vepsilon_t \mid \hat \vx_t=\hat \vx, \vu_t=\vu]$ be the conditional process noise means for the true and candidate parameterizations, respectively.
In the intervention-free regime $\vu=0$, comparing conditional means under affine identifiability gives
\begin{equation}
    g_{\hat\phi}(\mM\vx+\vc)+\hat\vb
    =\mM g_\phi(\vx)+\vc+\mM\vb.
    \label{eq:step3_baseline_mean}
\end{equation}
The above equality initially holds almost everywhere under the intervention-free state distribution. By \Asmref{as:full_latent_support} and \Asmref{as:non_deg_dyn2}, the latent state distribution has full support and both sides are continuous, making the equality hold everywhere in $\vx\in\R^D$.

Since the noise means are constant in $\vx$ and the intervention (\Appref{app:proof_setup}), differentiating with respect to $\vx$ yields
\begin{equation}
    \mJ_{g_{\hat\phi}}(\mM\vx+\vc)\mM
    =\mM\mJ_{g_\phi}(\vx).
    \label{eq:step3_baseline_jacobian}
\end{equation}

For a fixed intervention input $\vu$, define
$\mD_{\vu}:=\operatorname{diag}(\vm(\vu))$ and
$\hat\mD_{\vu}:=\operatorname{diag}(\hat\vm(\vu))$.
Comparing conditional means under this input gives
\begin{equation}
    \mM\mD_{\vu}g_\phi(\vx)+\mM\mB\vu+\vc+\mM\vb
    =\hat\mD_{\vu}g_{\hat\phi}(\mM\vx+\vc)+\hat\mB\vu+\hat\vb.
    \label{eq:step3_intervention_mean}
\end{equation}
Again, full-support and continuity assumptions ensure this holds everywhere in $\vx\in\R^D$. Differentiating with respect to $\vx$ yields
\begin{equation}
    \mM\mD_{\vu}\mJ_{g_\phi}(\vx)
    =\hat\mD_{\vu}\mJ_{g_{\hat\phi}}(\mM\vx+\vc)\mM.
\end{equation}

Substituting Eq.~\eqref{eq:step3_baseline_jacobian} into the right-hand side gives
\begin{equation}
    \mM\mD_{\vu}\mJ_{g_\phi}(\vx)
    =\hat\mD_{\vu}\mM\mJ_{g_\phi}(\vx)
\end{equation}
By \Asmref{as:non_deg_dyn2}, $\mJ_{g_\phi}(\vx^*)$ is invertible at some $\vx^*\in\R^D$, thus we get
\begin{equation}
    \mM\mD_{\vu}=\hat\mD_{\vu}\mM.
    \label{eq:step3_mask_relation}
\end{equation}
Entrywise, this means $(m_j(\vu)-\hat m_i(\vu))M_{ij}=0$. Hence, whenever candidate coordinate $i$ and true coordinate $j$ have different intervention statuses, we have $M_{ij}=0$. Consequently,
\begin{equation}
    \mM_{\mathcal{J}',-\mathcal{J}}=0, \qquad
    \mM_{-\mathcal{J}',\mathcal{J}}=0.
\end{equation}
Moreover, Eq.~\eqref{eq:step3_mask_relation} gives
$\hat\mD_{\vu}=\mM\mD_{\vu}\mM^{-1}$, so the masks have equal rank and $|\mathcal{J}'|=|\mathcal{J}|$. After aligning the target sets, the diagonal blocks of $\mM$ are therefore square and invertible, since $\mM$ is invertible.
\end{proof}

\paragraph{From block to full identifiability.}
Under the unordered pairs condition (\Asmref{as:unordered_pairs}), the above argument applies to every pair of latent coordinates, forcing $\mM$ to be diagonal up to permutation. Hence $\mM = \mLambda \mPi$, where $\mPi$ is a permutation matrix and $\mLambda$ is diagonal and invertible, yielding identifiability up to permutation, scaling, and shift as shown by \citet{nejat_wang_2025}.

\subsection{Nonlinear input maps}
\label{app:nonlinear_inputs}

\Corref{cor:perm_scale_shift} also holds when the linear input map $\mB\vu_t$ is replaced by an arbitrary deterministic map $h(\vu_t)$, provided there exists an intervention-free regime on the support of $\vu$.

\begin{proposition}[Nonlinear input maps]
\label{prop:nonlinear_inputs}
Replace $\mB\vu_t$ in Eq.~\eqref{eq:general_issm_x} by an arbitrary deterministic map $h:\R^M\to\R^D$, and let $m_{t,j} := \ind\{h_j(\vu_t)=0\}$. Assume the conditions of \Corref{cor:perm_scale_shift}, with \Asmref{as:unordered_pairs} applied to the coordinates that receive nonzero input. If there exists an intervention-free regime $\vu^0\in\operatorname{supp}(\vu)$ such that $h(\vu^0)=0$ and $\hat h(\vu^0)=0$, then the latent variables remain identifiable up to permutation, scaling, and shift.
\end{proposition}

\begin{proof}[Proof of \Propref{prop:nonlinear_inputs}]
    Here $m_{t,j}=\ind\{h_j(\vu_t)=0\}$. We verify that the three-step proof of \Corref{cor:perm_scale_shift} remains valid when replacing $\mB\vu$ by $h(\vu)$ with one additional assumption on the input maps $h$ and $\hat h$, needed for \emph{steps 2 and 3}.
    
    \emph{Step 1} is unchanged because it does not involve the input map.
    
    \emph{Step 2} requires an intervention-free regime for both parameterizations because the transition Jacobian $\operatorname{diag}(\vm_t)\mJ_{g_\phi}(\vx_t)$ is singular during active interventions. By assumption, $h(\vu^0)=\hat h(\vu^0)=0$, so both masks equal $\vone$ at $\vu_t=\vu^0$. With this, the latent transitions become $\vx_{t+1}=g_\phi(\vx_t)+\vepsilon_t$ and $\hat\vx_{t+1}=g_{\hat\phi}(\hat\vx_t)+\hat\vepsilon_t$, and \Thmref{thm:affine_identifiability} can be applied unchanged. The resulting affine map $\psi(\vx)=\mM\vx+\vc$ is independent of the intervention and therefore holds both at $\vu^0$ and under active interventions.

    \emph{Step 3} uses $\vu^0$ as the intervention-free regime for both parameterizations. For any fixed input $\vu$, the terms $h(\vu)$ and $\hat h(\vu)$, just like $\mB\vu$ and $\hat\mB\vu$, disappear upon differentiation with respect to $\vx$. The remainder of the proof in \Appref{app:step3} is unchanged.
    \end{proof}

\begin{remark}
\label{rem:input_map_support}
Note that the data constrains $h$ only on $\operatorname{supp}(\vu)$. If the observed interventions comprise finitely many channel-amplitude combinations, one-hot encoding these combinations allows a linear input map with one column per combination to represent any $h$ exactly on the observed support.
\end{remark}

\subsection{Observation masking}
\label{app:obs_masking}

In the human iEEG study, we mask out observation channels that correspond to active stimulation sites on each trial to exclude saturation artefacts. We show here that the identifiability guarantees above remain valid in this setting with modest additional conditions.

\begin{definition}[Exogenous observation masking]
\label{def:obs_masking}
Let $\vo \in \{0,1\}^N$ denote a binary mask indicating which observation coordinates are retained. On trial $k$, a mask $\vo_k$ applies. The mask is known and determined only by observed exogenous variables; the observation noise is independent of these variables and the latent states taken jointly. Write $\mathcal{O}(\vo) := \{n : o_n = 1\}$ for the indices of the observed coordinates and let $\mS_{\vo} : \R^N \to \R^{|\mathcal{O}(\vo)|}$ denote projection onto those coordinates; we write $\mS_{\mathcal{A}}$ analogously for projection onto an index set $\mathcal{A}\subseteq\{1,\dots,N\}$. Finally, define the masked noiseless signal as $\ybar_{t,\mathcal{O}(\vo)} := \mS_{\vo} f_\theta(\vx_t)$, and masked observations as $\vy_{t,\mathcal{O}(\vo)} := \mS_{\vo} \vy_t$.
\end{definition}

We verify that the three-step proof of \Corref{cor:perm_scale_shift} remains valid under \Defref{def:obs_masking}. For each step, we state what the masking changes, give sufficient conditions under which the step still holds, and discuss plausibility in the human iEEG analysis.

\paragraph{Step 1: bounded completeness under observation masking.}
Step~1 (\Appref{app:step1}) requires bounded completeness (\Asmref{as:bounded_completeness}) for the observation model. Under masking, the analogous requirement is that $P(\vy_{t,\mathcal{O}(\vo)} \mid \ybar_{t,\mathcal{O}(\vo)})$ is bounded complete in the sense of \Asmref{as:bounded_completeness}. Below we provide a sufficient condition that applies to the human iEEG analysis.

\begin{proposition}[Bounded completeness under observation masking]
\label{prop:masked_completeness}
Suppose that:
\begin{enumerate}[label=(\roman*), nosep]
    \item observations are generated by additive noise $\vy_t = \ybar_t + \veta_t$, where $\ybar_t$ is the noiseless signal and $\veta_t \indep \ybar_t$;
    \item the noise $\veta_t$ admits a density $p_\veta$ on $\R^N$; and
    \item the characteristic function $\varphi_\veta$ of $\veta_t$ satisfies $\varphi_\veta(\vs) \neq 0$ for all $\vs \in \R^N$.
\end{enumerate}
Then the conditional distribution $P(\vy_t \mid \ybar_t)$ satisfies \Asmref{as:bounded_completeness}. The same conclusion holds under any mask satisfying \Defref{def:obs_masking}.
\end{proposition}

\begin{proof}

We first prove the claim for the unmasked case, and later we show that the argument also holds under masking.

By (i) and (ii), the conditional density of $\vy_t$ given $\ybar_t$ is $p_\veta(\vy_t - \ybar_t)$.
Let $\mu$ be a finite signed measure satisfying
\begin{equation*}
    \int P(\vy_t\in E\mid\ybar_t=\vz)\,d\mu(\vz)=0
    \qquad\text{for every measurable }E.
\end{equation*}
Using the conditional density and interchanging the integrals gives
$\int_E (p_\veta * \mu)(\vy)\,d\vy=0$ for every such $E$. Since $p_\veta\in L^1$ and $\mu$ is finite, the convolution is integrable, so
\begin{equation}
    (p_\veta * \mu)(\vy)=\int p_\veta(\vy-\vz)\,d\mu(\vz)=0
    \quad\text{for Lebesgue-almost every }\vy.
    \label{eq:bounded_completeness_condition_part1}
\end{equation}
Taking Fourier transforms in $\vy$, with $\mathcal{F}\mu(\vs) := \int e^{i \vs^\top \vz}\, d\mu(\vz)$, gives
\begin{equation}
    \mathcal{F}\mu(\vs)\, \varphi_\veta(\vs) = 0 \quad \text{for all } \vs.
\end{equation}

By (iii), $\varphi_\veta(\vs) \neq 0$ for all $\vs$, thus $\mathcal{F}\mu(\vs) = 0$ for all $\vs$. Uniqueness of the Fourier--Stieltjes transform of finite signed measures then implies $\mu = 0$, as required.

Now we show that the conditions are preserved under masking. By \Defref{def:obs_masking}, the mask is exogenous and fixed, so the additive noise model and independence of~(i) are preserved:
\begin{equation}
    \vy_{t,\mathcal{O}(\vo)} = \mS_{\vo} \ybar_t + \mS_{\vo} \veta_t 
    \quad \text{and} \quad
    \veta_{t,\mathcal{O}(\vo)} \indep \ybar_{t,\mathcal{O}(\vo)}.
\end{equation}

Condition~(ii) is preserved because the marginal of a density is a density, i.e., integrating $p_\veta$ over the masked-out coordinates yields a density $p_{\veta_{\vo}}$ of $\mS_{\vo} \veta_t$ on $\R^{|\mathcal{O}(\vo)|}$.

Finally, the characteristic function of the marginal noise satisfies
\begin{equation}
    \varphi_{\veta_{\vo}}(\vs_{\vo}) = \varphi_\veta\!\left(\mS_{\vo}^\top \vs_{\vo}\right),
\end{equation}

so a non-vanishing $\varphi_\veta$ implies $\varphi_{\veta_{\vo}}(\vs_{\vo}) \neq 0$ for all $\vs_{\vo}$. Hence, the above convolution argument shown for the unmasked case also applies to the masked case.
\end{proof}

\begin{remark}
In the human iEEG analysis, the mask is fully determined by whichever stimulation site is active during that trial, so \Defref{def:obs_masking} applies. Additionally, we assume Gaussian observation noise, $\vy_t = \ybar_t + \veta_t$ with $\veta_t \sim \mathcal{N}(0, \mSigma_\veta)$ and $\mSigma_\veta$ positive definite. Gaussian noise is additive and its characteristic function never vanishes. The non-singular covariance ensures that $\veta_t$ admits a density on $\R^N$. Thus, \Propref{prop:masked_completeness} implies that Step~1 holds under masking in this application.
\end{remark}

\paragraph{Step~2: affine identifiability under observation masking.}
Observation masking may break \Asmref{as:regularity} of Step~2 (\Appref{app:step2}), which requires the emission function to be injective and continuously differentiable, with continuously differentiable inverse on its image: projecting to $\mathcal{O}(\vo)$ can break injectivity even when the full decoder $f_\theta$ is injective. Below we give sufficient conditions under which affine identifiability is preserved, then justify their plausibility in our setting.

\begin{proposition}[Affine identifiability under observation masking]
\label{prop:masked_regularity}
Assume that all conditions of \Thmref{thm:affine_identifiability} hold, and that for every mask the data contain transitions without active interventions ($\vm_t = \vone$). Assume additionally that:
\begin{enumerate}[label=(\roman*), nosep]
    \item \label{item:masked_reg_single} for every mask $\vo$, the restricted decoders $\mS_{\vo} \circ f_\theta$ and $\mS_{\vo} \circ f_{\hat\theta}$ are injective and continuously differentiable, with continuously differentiable inverses on their images;
    \item \label{item:masked_reg_pair} for every pair of masks $\vo, \vo'$, the restricted decoders $\mS_{\mathcal{O}(\vo) \cap \mathcal{O}(\vo')} \circ f_\theta$ and $\mS_{\mathcal{O}(\vo) \cap \mathcal{O}(\vo')} \circ f_{\hat\theta}$ are injective.
\end{enumerate}
Then affine identifiability is preserved under observation masking in the sense of \Defref{def:obs_masking}.
\end{proposition}

\begin{proof}
Fix two masks $\vo$ and $\vo'$. By~\ref{item:masked_reg_single}, we can apply Step~2 separately under each mask, using its transitions without active interventions, yielding affine maps $\psi_{\vo}(\vx) = \mM_{\vo} \vx + \vc_{\vo}$ and $\psi_{\vo'}(\vx) = \mM_{\vo'} \vx + \vc_{\vo'}$ such that
\begin{equation}
    \mS_{\vo} f_\theta(\vx) = \mS_{\vo} f_{\hat\theta}(\psi_{\vo}(\vx)),
    \qquad
    \mS_{\vo'} f_\theta(\vx) = \mS_{\vo'} f_{\hat\theta}(\psi_{\vo'}(\vx)).
\end{equation}
Restricting both identities to $\mathcal{O}(\vo) \cap \mathcal{O}(\vo')$ gives
\begin{equation}
    \mS_{\mathcal{O}(\vo) \cap \mathcal{O}(\vo')} f_\theta(\vx)
    =
    \mS_{\mathcal{O}(\vo) \cap \mathcal{O}(\vo')} f_{\hat\theta}(\psi_{\vo}(\vx))
    =
    \mS_{\mathcal{O}(\vo) \cap \mathcal{O}(\vo')} f_{\hat\theta}(\psi_{\vo'}(\vx)).
\end{equation}
By~\ref{item:masked_reg_pair}, injectivity of $\mS_{\mathcal{O}(\vo) \cap \mathcal{O}(\vo')} \circ f_{\hat\theta}$ implies $\psi_{\vo}(\vx) = \psi_{\vo'}(\vx)$ for all $\vx \in \R^D$, which is the latent support under both masks by \Asmref{as:gaussian_latents}. Hence $\mM_{\vo} = \mM_{\vo'}$ and $\vc_{\vo} = \vc_{\vo'}$. Since $\vo$ and $\vo'$ were arbitrary, a single affine map $\hat \vx = \mM \vx + \vc$ holds across all mask patterns.
\end{proof}

\begin{remark}
In the human iEEG analysis, each mask removes only the two channels at the active stimulation site (at most four under pairwise overlap, out of $N=99$ outputs). For $\mS_{\vo} \circ f_\theta$ to fail injectivity, distinct latents $\vx \neq \vx'$ would have to agree on all unmasked channels while differing only on masked ones. That is, the ${\sim}4$ masked coordinates would have to contain unique information needed to distinguish these latents that is absent from the remaining ${\sim}95$ channels. A similar argument applies to the continuously differentiable inverse in~\ref{item:masked_reg_single}: it requires the Jacobian $\mS_{\vo} \mJ_{f_\theta}(\vx)$ to keep full rank $D$ at every $\vx$, which fails only if, at some $\vx$, a small change of the latent state along some direction affects the masked channels alone. The masked channels are the ones closest to the stimulated tissue; it is therefore not unthinkable that they contain identifying information that other channels lack. However, other contacts on the same electrode shaft remain observed, and stimulation also evokes responses at distant channels. We therefore consider such unique information in the few masked channels unlikely and treat conditions~\ref{item:masked_reg_single} and~\ref{item:masked_reg_pair} as a mild robustness assumption. Finally, each trial contains a single stimulation pulse, so all other time steps are free of active interventions, as required by \Propref{prop:masked_regularity}.
\end{remark}

\paragraph{Step~3: refinement under interventions.}
Step~3 (\Appref{app:step3}) concerns only the latent transition under hard interventions and does not involve the observation model. What is relevant in this context is that Step~3 combines constraints obtained from different intervention channels to restrict a single affine transformation $\mM$. The conclusion of \Propref{prop:masked_regularity} that the affine maps obtained under different observation masks must coincide, is therefore essential. However, once Steps~1 and~2 establish a common affine map $\hat \vx = \mM\vx+\vc$ across masks, Step~3 remains valid and requires no additional assumptions.

\section{Variational Inference}
\label{app:inference}

We perform parameter learning and latent state inference using variational inference. Given observations $\vy_{1:T}$ and interventions $\vu_{1:T}$, our goal is to infer the posterior distribution over latent trajectories $p_{\Theta}(\vx_{1:T} \mid \vy_{1:T}, \vu_{1:T})$ while learning model parameters $\Theta = (\theta, \phi, \mB, \mQ)$.

Since exact inference is intractable due to nonlinearities in both the transition and emission models, we introduce an approximate posterior $q_{\Phi}(\vx_{1:T} \mid \vy_{1:T}, \vu_{1:T})$ and optimize the evidence lower bound (ELBO):
\begin{align}
\mathcal{L}(\Phi, \Theta) = \E_{q_{\Phi}} \left[ \log p_{\Theta}(\vy_{1:T}, \vx_{1:T} \mid \vu_{1:T}) - \log q_{\Phi}(\vx_{1:T} \mid \vy_{1:T}, \vu_{1:T}) \right].
\end{align}

We adopt amortized variational inference and parameterize the approximate posterior using a recurrent neural network. Specifically, we use an LSTM-based recognition model to produce the mean and variance of a Gaussian approximate posterior over latent states at each time step. Latent variables are reparameterized as $\vx_t = \vmu_t + \vsigma_t \odot \vxi_t$, where $\vxi_t \sim \mathcal{N}(0, \mI)$.

To match the interventional structure from Eq.~\eqref{eq:general_issm_x}, we apply the mask $\vm_t$ to the approximate posterior mean during training:
\begin{align}
\vmu_t \leftarrow \vm_t \odot \vmu_t + \mB \vu_t.
\end{align}

The model parameters and variational parameters are jointly optimized using stochastic gradient descent on the negative ELBO.

\section{Statistical Analysis}
\label{app:stats}

Unless stated otherwise, each model is trained with 10 random seeds.
Reported means are accompanied by the standard error of the mean (s.e.m.).
When reporting per-trial performance in real-data experiments, we first average predictions across seeds for each held-out trial, then report the mean and s.e.m. across trials within each evaluation tier (e.g., control vs. stimulation). 
Otherwise, reported accuracy metrics are mean $\pm$ s.e.m. across seeds.
Model comparisons use a two-sided paired Wilcoxon signed-rank test on seed-level scores ($n=10$). Where applicable, $p$-values are Benjamini-Hochberg FDR-corrected for multiple comparisons.

\section{FitzHugh-Nagumo Experiment Details}
\label{app:fhn_details}

We provide full details for the simulation experiments in \Secref{sec:fhn}.

\subsection{Data Generation}
\label{app:fhn_data}

We simulate the two-dimensional FHN system
\begin{align}
    \dot{v} &= v - \tfrac{v^3}{3} - w + I_\mathrm{ext}, &
    \dot{w} &= \tfrac{v + a - bw}{\tau},
\end{align}
with parameters $a{=}0.7$, $b{=}0.8$, $\tau{=}12.5$, $I_\mathrm{ext}{=}0.5$ (oscillatory regime; true latent dimension $D{=}2$).
Trajectories have length $T{=}50\,\mathrm{s}$ at $\Delta t{=}0.1\,\mathrm{s}$, preceded by a 5-second burn-in that is discarded before training.
We use 200 training trajectories (batch size 25) and 10 validation trajectories.
Initial conditions are sampled as $\vx_0 \sim \mathcal{N}(0,\, \sigma_0^2 \mI_D)$ with $\sigma_0{=}0.05$ before the burn-in.
Latent dynamics include additive Gaussian process noise with $\sigma_x{=}0.05$; observations are noiseless ($\sigma_y{=}0$).

\paragraph{Observation model.}
Observations are obtained via $\vy_t = \tanh(\mC \vx_t)$, where $\mC \in \R^{N \times D}$ is a fixed random matrix with orthonormal columns ($N{=}4$ observations).

\paragraph{Interventions.}
Interventions are delivered on $M{=}4$ input channels via random (Poisson) pulse stimulation: pulses arrive as a Poisson process at rate $\lambda{=}0.20\,\mathrm{s}^{-1}$, each assigned to a uniformly random input channel, with pulse duration $0.25$~s, amplitude drawn uniformly from $[-2, 2]$, and additive Gaussian noise $\sigma{=}0.25$. Of the training trajectories, 10\% are non-interventional (control trials) with no stimulation applied.

\subsection{Model Architecture}
\label{app:toy_issm_arch}

Linear and nonlinear variants share the same emission and recognition architectures and differ only in the dynamics parameterization.

\paragraph{Dynamics.}
The nonlinear model uses $g_\phi(\vx_t) = \mathrm{MLP}(\vx_t)$, implemented as a two-hidden-layer network with 8 units per layer and GeLU activations. The linear model uses $g_\phi(\vx_t) = \mA\vx_t$, with $\mA\in\R^{D\times D}$ learned.
Both share a learned input matrix $\mB\in\R^{D\times M}$, initialized to the identity (zero-padded as needed) and subject to the $\mB$-sparsity mechanisms below.

\paragraph{$\mB$-sparsity mechanisms.}
Two mechanisms enforce sparsity on $\mB$.
First, at each forward pass we treat entries with $|B_{ij}|<\tau_B$ as inactive when computing the hard-intervention mask ($\tau_B{=}10^{-3}$ fixed for all experiments), so tiny numerical residuals do not trigger spurious interventions.
Second, during training we hard-threshold $\mB$ elementwise after each optimizer step (outside backpropagation): entries with $|B_{ij}|<\tau_t$ are set to zero, where $\tau_t$ increases linearly from $0$ to $\tau_\mathrm{max}{=}0.10$ over the first $p_\tau{=}0.10$ fraction of the maximum number of epochs and is fixed thereafter.
We use hard thresholding rather than an $\ell_1$ penalty so that surviving entries are not shrunk toward zero while small residual entries are set to exact zero; dense support on $\mB$ also violates the unordered-pairs condition needed for \Corref{cor:perm_scale_shift}.
The warmup avoids premature pruning early in training.
On FHN (nonlinear iSSM, 10 seeds), disabling the schedule ($\tau_\mathrm{max}{=}0$) reduced latent accuracy from $0.97 \pm 0.02$ to $0.74 \pm 0.04$ and degraded $\mB$ recovery (paired Wilcoxon signed-rank test: both $p{=}0.004$); although a milder threshold $\tau_\mathrm{max}{=}0.05$ already matched the default model's performance.

\paragraph{Emission.}
The emission model factorises into a linear readout and a nonlinear residual correction:
\begin{equation}
    f_\theta(\vx_t) = \mW_{\mathrm{skip}}\, \vx_t + \mathrm{MLP}_\theta(\vx_t),
\end{equation}
where $\mW_{\mathrm{skip}}\in\R^{N\times D}$ is a learned linear readout and the nonlinear branch projects $\vx_t$ to $H_\mathrm{em}{=}32$ dimensions (linear + GeLU), passes it through $n_{\mathrm{blocks}}{=}2$ pre-norm residual blocks (LayerNorm $\to$ Linear $\to$ GeLU $\to$ Linear with skip connection), and projects back to $N$. The nonlinear branch is initialised with near-zero output weights so the model starts approximately linear.

\paragraph{Recognition.}
The approximate posterior $q_\Phi(\vx_{1:T}\mid \vy_{1:T}, \vu_{1:T})$ is parameterized by a shared-backbone LSTM: inputs $(\vy_t, \vu_t)$ are linearly projected to $H_\mathrm{rec}{=}32$ dimensions (LayerNorm + GeLU), processed by a single-layer LSTM, and decoded by two separate heads for the posterior mean and log-variance.

\subsection{Training}
\label{app:toy_issm_training}

Models are trained to maximize the ELBO (standard $\beta{=}1$ VAE objective) using Adam with learning rate $3\times10^{-3}$, gradient clipping (max norm $1.0$), and $\ell_2$ weight decay $10^{-3}$. The KL term is linearly warmed up over the first $p_\beta{=}0.20$ fraction of epochs; the training-time $\mB$-sparsity warmup follows the schedule above. Early stopping monitors rollout MAE on the validation set with patience 50 epochs, up to a maximum of 1000 epochs.

\subsection{Evaluation Protocol}
\label{app:toy_eval}

All toy-model results in \secrefs{sec:fhn}{sec:dmm}{sec:stim} use the same evaluation protocol.
After training, we score each seed on 100 newly simulated test trajectories from the true dynamical system. 
Using the training $\Delta t$ and burn-in $T_\mathrm{burn}$, we simulate $T_\mathrm{burn} + T_\mathrm{encode} + T$ seconds per trial ($T_\mathrm{encode}{=}5\,\mathrm{s}$, $T{=}100\,\mathrm{s}$) and discard the burn-in.

\paragraph{Test interventions.}
Every test trial receives Poisson pulse stimulation.
Pulse duration, input noise, and the amplitude interval are inherited from the training stimulator, with the amplitude range scaled by $1.2$ to probe interventions stronger than those seen during training.
The 100 test trials are distributed approximately evenly over $7$ stimulation rates uniformly spaced in $[0.2, 0.8]\,\mathrm{s}^{-1}$.

\paragraph{Trial structure and scoring.}
Reconstruction, rollout, and latent accuracy (\Secref{sec:metrics}) are computed on the $100\,\mathrm{s}$ evaluation segment (after the encode window).
We compute one Pearson correlation per trial then average over trials within each seed.
Correlations use only non-interventional timesteps where all input channels are zero.
Trials on which the model produces non-finite outputs (e.g.\ a diverged rollout) score zero.

\paragraph{Alignment.}
For latent accuracy we permutation-align inferred latents by Hungarian matching on the inferred and true input matrices~$\mB$, then fit a per-dimension affine map (scale and shift) from the driven full-trial posterior at non-interventional timesteps.

\subsection{Computational Resources}
\label{app:fhn_compute}

All FHN experiments were run on CPU with 2\,GB RAM. Per seed, training takes approximately 30-45 min (variation due to early stopping).

\section{Van Der Pol Oscillator Experiments}
\label{app:vdp}

\begin{figure}[t]
    \centering
    \includegraphics[width=\linewidth]{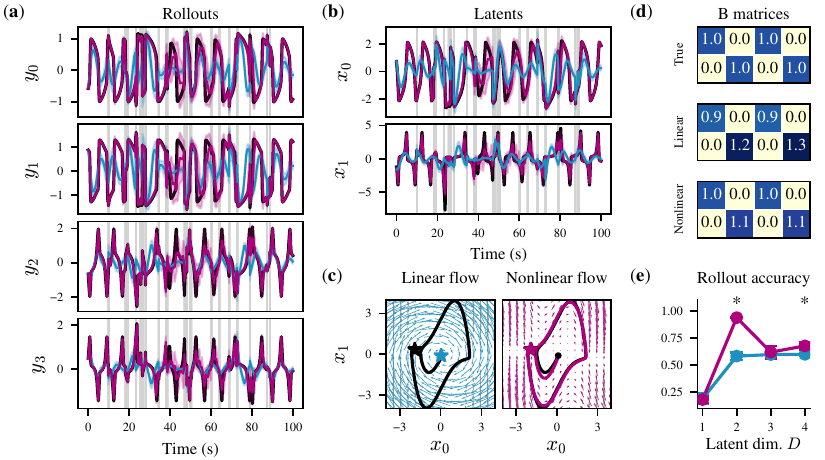}
    \caption{\textbf{Nonlinear iSSM identifies Van Der Pol oscillator latent dynamics.}
    (a,b)~Open-loop rollout and aligned latent trajectories on a test trial: true (black), linear iSSM (blue), nonlinear iSSM (magenta); mean $\pm$ s.d.\ across 10 seeds. Gray lines mark interventions.
    (c)~Latent flow fields from the seed with highest Jacobian similarity to ground truth. Gray/black: true flow (arrows) and trajectory (line); blue/magenta: inferred linear/nonlinear flow and trajectories.
    (d)~Median inferred input map $\mB$ after permutation-scale alignment.
    (e)~Latent-dimension sweep over $D \in \{1,2,3,4\}$ shows that model selection picks out the true dimension ($D=2$).
    }
    \label{fig:vanderpol_toymodels}
  \end{figure}

The Van der Pol (VdP) oscillator with its square-like relaxation limit cycle has qualitatively different attractor geometry compared to FHN, but it also lies outside the linear iSSM hypothesis class.

We simulated VdP dynamics
\begin{align}
    \dot{x}_1 &= x_2, &
    \dot{x}_2 &= \mu(1-x_1^2)\,x_2 - x_1,
\end{align}
with $\mu = 2$ ($D=2$ latent dimensions).
Model architecture and training follow \Appsref{app:toy_issm_arch}{app:toy_issm_training}.
Data generation follows the same protocol as \Appref{app:fhn_data}, with the following deviations:
$\Delta t{=}0.05\,\mathrm{s}$;
process noise $\sigma_x{=}0.002$;
observation nonlinearity $\operatorname{arsinh}$ instead of $\tanh$;
stimulation rate $\lambda{=}0.45\,\mathrm{s}^{-1}$;
recognition hidden dimension $H_\mathrm{rec}{=}64$;
learning rate $2\times10^{-3}$;
maximum epochs 2000;
and $\mB$-sparsity warmup fraction $p_\tau{=}0.05$.

\paragraph{Observation nonlinearity.}
We use $\operatorname{arsinh}$ rather than $\tanh$ as the pointwise observation nonlinearity. The VdP oscillator at $\mu=2$ produces oscillations with a substantially larger amplitude than FHN, and the random injective projection $\mC$ can amplify this further. At such value ranges, $\tanh$ saturates and produces near-zero gradients at the extremes, creating an optimization challenge that is unrelated to identifiability. The $\operatorname{arsinh}$ nonlinearity is unbounded (growing as $\log(2|x|)$ for large $|x|$), preserves injectivity, and avoids gradient saturation.

The VdP experiments replicate the FHN trends: The nonlinear iSSM again outperforms its linear counterpart (\Figref{fig:vanderpol_toymodels}), achieving rollout accuracy of $r = 0.94 \pm 0.01$ s.e.m. versus $0.58 \pm 0.04$ and latent accuracy of $r = 0.93 \pm 0.01$ versus $0.55 \pm 0.03$ (\Figref{fig:vanderpol_toymodels}a,b; Wilcoxon signed-rank test: both $p=0.0020$). The best nonlinear model recovers the characteristic square-like VdP attractor, whereas the best linear model produces a spiral attractor (\Figref{fig:vanderpol_toymodels}c). Nonlinear iSSM also more faithfully recovers $\mB$ (\Figref{fig:vanderpol_toymodels}d). Model selection picks out the true dimension ($D=2$) for both models (\Figref{fig:vanderpol_toymodels}e).

\section{Decision-Making Model Experiment Details}
\label{app:dmm_details}

We provide full details for the simulation experiments in \Secref{sec:dmm}.

\subsection{Dynamical Systems}
\label{app:dmm_systems}

We simulate three two-dimensional systems ($D=2$) as follows.

\paragraph{Bistable well (nonlinear integration).}
\begin{align}
    \dot{x}_1 &= a x_1^3 + b x_1 + c_\mathrm{oh}, &
    \dot{x}_2 &= c x_2 + c_\mathrm{oh},
\end{align}
with $a{=}-0.6$, $b{=}2.0$, $c{=}-1.0$, and $c_\mathrm{oh}{=}0.1$.
Evidence is integrated via an unstable saddle separating two stable choice states.

\paragraph{Line attractor (perfect integration).}
\begin{equation}
    \dot{\vx} = \mA \vx,
\end{equation}
where $\mA = \mathrm{Rot}(\gamma)\, \mR \mLambda \mR^{-1}$ has one marginally stable direction (eigenvalue $0$) and one stable direction (eigenvalue $\approx -1.21$ for the values below).
Here $\mLambda=\mathrm{diag}(0,\lambda_1)$, $\mR=[\vr_0,\vl_0]^\top$ has rows $\vr_0^\top$ and $\vl_0^\top$, and $\mathrm{Rot}(\gamma)$ is a $2\times2$ rotation matrix.
We use $\vl_0=[1,1]^\top$, $\vr_0=[1,0]^\top$ (each normalized), $\lambda_1{=}-1.0$, and $\gamma{=}45^\circ$.

\paragraph{Point attractor (input-driven dynamics).}
\begin{equation}
    \dot{\vx} = \mathrm{diag}(a_1,a_2)\,\vx,
\end{equation}
with $a_1{=}-0.5$ and $a_2{=}-1.0$.
Both latent dimensions decay to a fixed point at the origin, so decision-related activity is inherited from inputs rather than generated locally.

\subsection{Data Generation}
\label{app:dmm_data}

The three systems share the same observation and intervention protocol.
Trajectories have length $T{=}20\,\mathrm{s}$ at $\Delta t{=}0.1\,\mathrm{s}$ with no burn-in.
We generate 40 training, 10 validation, and 100 test trajectories (batch size 20).
All systems use $D{=}2$ latent dimensions, $N{=}10$ observations, and $M{=}2$ input channels.
Latent dynamics include additive Gaussian process noise with $\sigma_x{=}0.1$; observations are noiseless ($\sigma_y{=}0$).

\paragraph{Observation model.}
Observations are obtained via $\vy_t = \tanh(\mC \vx_t)$, where $\mC \in \R^{N \times D}$ is a fixed random matrix with orthonormal columns.

\paragraph{Interventions.}
Interventions are delivered on $M{=}2$ input channels via random pulse stimulation: pulses arrive as a Poisson process at rate $\lambda{=}0.75\,\mathrm{s}^{-1}$, each assigned to a uniformly random input channel, with pulse duration $0.25$~s, amplitude drawn uniformly from $[-3, 3]$, and additive Gaussian noise $\sigma{=}0.25$. All training trajectories contain interventions.

\subsection{Training}
\label{app:dmm_training}

We fit linear and nonlinear iSSMs to interventional data from each system.
Model architecture and training follow \Appsref{app:toy_issm_arch}{app:toy_issm_training}, with learning rate $5\times10^{-3}$ and batch size 20.
Linear and nonlinear variants differ only in the dynamics parameterization.
We train 10 random seeds per model class and evaluate on held-out trajectories with novel interventions.

\subsection{Computational Resources}
\label{app:dmm_compute}

All decision-making experiments were run on CPU with 2\,GB RAM. Per seed, training takes approximately 15--30 minutes (variation due to early stopping).

\section{Macaque dlPFC Dataset Experiment Details}
\label{app:kiani_details}

\subsection{Data}
\label{app:kiani_data}

We use spike-count recordings from macaque prefrontal cortex (area 8Ar) collected during resting-state micro-stimulation experiments~\citep{Nejatbakhsh2023PredictingTE}.
Data were recorded at $f_s{=}30$~Hz across $N{=}96$ channels.
Trials are extracted by windowing $\pm$1\,s around each stimulation event (stimulation trials) or around randomly sampled time points matched in count (control trials), giving $T{=}60$ bins per trial at $\Delta t{=}33$~ms. Both conditions are used during training.

\paragraph{Input construction.}
The stimulation input $\vu_t \in \R^M$ is a binary indicator that is one at the stimulated electrode's channel during stimulation and zero elsewhere.
We compress $\vu$ to only the $M{=}12$ electrodes that were ever stimulated across all sessions, so that each entry of $\vu$ has a direct causal interpretation.
Control trials have $\vu_t = 0$ throughout.
Train, validation, and test sets are split with stratification by trial type (stimulation vs.\ control) to ensure both conditions are represented in each split; 10 trials are reserved for validation and 60 for testing.

\subsection{Model Architecture}
\label{app:kiani_arch}

The linear and nonlinear models use the same implementation as the simulation experiments (\Appref{app:toy_issm_arch}), with the dataset-specific modifications below.
They differ only in their dynamics parameterization.

\paragraph{Likelihood.}
Because observations are spike counts, we use a Poisson likelihood: $p(\vy_t \mid \vx_t) = \mathrm{Poisson}\!\left(\exp(f_\theta(\vx_t))\right)$ applied independently per channel, where $f_\theta(\vx_t)$ outputs log-rates.

\begin{figure}[h]
    \centering
    \includegraphics[width=\linewidth]{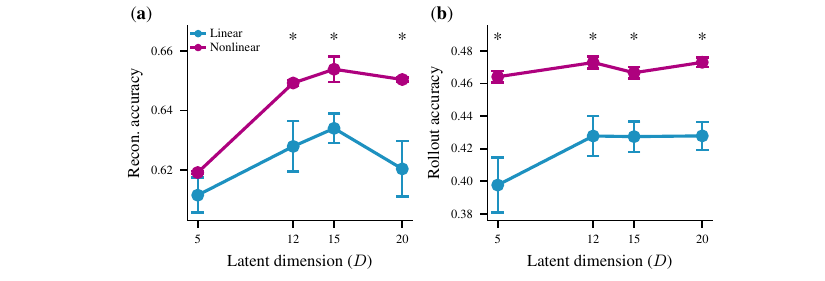}
    \caption{\textbf{Accuracy versus latent dimension $D$.}
    (a)~Reconstruction and (b)~rollout accuracy for linear (blue) and nonlinear (magenta) iSSMs; mean $\pm$ s.e.m.\ across 10 seeds.
    Asterisks mark significant differences ($p < 0.05$, FDR-corrected).}
    \label{fig:kiani_d_sweep}
  \end{figure}

\paragraph{Dynamics.}
The nonlinear model uses $g_\phi(\vx_t) = \mathrm{MLP}(\vx_t)$, implemented as a three-hidden-layer network with 64 units per layer and GeLU activations.
The linear model uses $g_\phi(\vx_t) = \mA\vx_t$, with $\mA\in\R^{D\times D}$ learned.
Both share a learned input matrix $\mB\in\R^{D\times M}$, initialized to an identity-like matrix with small perturbations and subject to the $\mB$-sparsity mechanisms described in \Appref{app:toy_issm_arch}.
We sweep the latent dimension over $D\in\{5,12,15,20\}$ and report $D{=}12$ in the main text.
The rollout advantage of the nonlinear iSSM holds at every $D$ (\Figref{fig:kiani_d_sweep}).

\paragraph{Emission.}
We use the residual-MLP emission described in \Appref{app:toy_issm_arch}, with hidden dimension $H_\mathrm{em}{=}32$, two residual blocks, and dropout probability $0.1$.

\paragraph{Recognition.}
The approximate posterior $q_\Phi(\vx_{1:T}\mid \vy_{1:T}, \vu_{1:T})$ is parameterized by a \emph{bidirectional} shared-backbone LSTM with hidden dimension $H_\mathrm{rec}{=}32$, input dropout probability $0.1$, and separate linear heads for the posterior mean and log-variance.
Because the recognition network is applied offline to the full trial, bidirectionality is appropriate and improves posterior quality.
During training, each neuron-level observation is independently masked with probability $0.1$.

\subsection{Training}
\label{app:kiani_training}

We use the same training loop and ELBO objective as in \Appref{app:toy_issm_training}.
We use Adam with learning rate $3\times10^{-3}$, batch size 25, $\ell_2$ weight decay $10^{-2}$, and gradient clipping at maximum norm $1.0$.
The KL term is warmed up linearly over the first $p_\beta{=}0.25$ fraction of epochs, while the training-time $\mB$-sparsity threshold increases to $\tau_\mathrm{max}{=}0.05$ over the first $p_\tau{=}0.50$ fraction.
Early stopping monitors rollout MAE on the validation set with patience 100 epochs, up to a maximum of 3000 epochs.
These settings are shared by the linear and nonlinear models.

\subsection{Evaluation}
\label{app:real_eval}

Reconstruction and rollout accuracy are computed on held-out test trials.
Reconstruction encodes the full trial with the bidirectional recognition network (\Appref{app:kiani_arch}).
For rollouts, the encoder sees only pre-stimulation observations and inputs: a fixed window of 31 timesteps ($\approx 1.0\,\mathrm{s}$ at 30~Hz), set to the earliest stimulation onset in the test set and shared across all trials including controls with $\vu_t = 0$ throughout.
$\vx_0$ is taken from the start of that window and the dynamics are integrated open-loop with the known test inputs over the full trial.
Rollout accuracy is the Pearson correlation between predicted and true observations, with the same channel masks as reconstruction.

\subsection{Unordered pairs estimate}
\label{app:kiani_up}

Permutation-scale-shift identifiability (\Corref{cor:perm_scale_shift}) requires the unordered pairs condition (\Asmref{as:unordered_pairs}).
To assess the degree to which the learned intervention design implements this condition, we read the $\tau_B$-thresholded support of the trained $\mB$ matrix ($\tau_B{=}10^{-3}$; \Appref{app:toy_issm_arch}).
Let $G_{dk} = \ind[|B_{dk}| \geq \tau_B]$ indicate whether an intervention via channel $k$ drives latent coordinate $d$.
Channel $k$ separates pair $(i,j)$ iff it drives exactly one of the two coordinates, i.e.\ $\mG_{i\cdot} \neq \mG_{j\cdot}$.
We report the fraction of separately intervenable pairs as:
\begin{equation}
\mathrm{up\_frac} = \frac{\#\{(i,j),\ i<j : \mG_{i\cdot} \neq \mG_{j\cdot}\}}{\binom{D}{2}},
\end{equation}
computed per seed and summarised as mean $\pm$ s.e.m. across seeds. 

For the main nonlinear macaque model ($D=12$), $\mathrm{up\_frac}=93.3 \pm 1.3\%$ (mean $\pm$ s.e.m. across 10 seeds). Note that this estimate is not a validation of the unordered pairs condition, as we have no ground-truth latent interventions.

\subsection{Computational Resources}
\label{app:kiani_compute}

All experiments were run on CPU with 8\,GB RAM. Training a single model run takes approximately 3--6 hours depending on early stopping; both the linear and nonlinear iSSM were trained with the same budget.

\section{Human iEEG Dataset Experiment Details}
\label{app:huang}

\subsection{Data}
\label{app:huang_data}

We use the human iEEG single-pulse electrical stimulation dataset of~\citet{huang2023electrical} (OpenNeuro \texttt{ds004457}) for subject \texttt{sub-2}.
Continuous SEEG was recorded at $f_s{=}2048$~Hz across 170 channels; biphasic single pulses were delivered to bipolar contact pairs at 4 or 6~mA.
Following the event filter of~\citet{huang2023electrical} (\texttt{status=good}), the preprocessed release contains 529 pulses at 60 bipolar stimulation sites.

\paragraph{Preprocessing.}
We replicate the signal-processing pipeline of~\citet{huang2023electrical}.
For each pulse we extract a $[-1, +2]$\,s window and apply, in order: (i) stimulation-artefact interpolation over $[-2, +15]$~ms (PCHIP spline fitted to surrounding clean samples), (ii) common-average reference over good non-stimulated channels, (iii) 60~Hz line-noise removal by spectrum fitting, and (iv) baseline correction (median over $[-0.5, -0.05]$\,s).

\paragraph{Channel and trial selection.}
We drop clinically bad channels, channels without anatomical location, white-matter contacts, and white-matter stimulation sites (and trials at removed sites), leaving $N{=}99$ recording channels, $M{=}50$ stimulation sites, and 437 trials.
Observations are decimated to 256~Hz with an anti-aliasing filter, cropped to $[-50, +200]$~ms around the pulse ($T{=}64$ bins at $\Delta t{=}3.9$~ms), and standardized per channel.

\paragraph{Input construction.}
The stimulation input $\vu_t \in \R^M$ is a single-sample impulse at pulse onset in the coordinate of the stimulated site, with amplitude equal to the delivered current (rescaled so the maximum maps to one) and zero elsewhere.

\paragraph{Masking and artefact trials.}
We exclude the two contacts of the stimulated pair from the likelihood for the whole trial (they saturate around the pulse).
\Appref{app:obs_masking} shows that identifiability is preserved under this masking, given modest additional conditions.
The same whole-trial stim-contact masks are used when scoring validation and test trials; masked channels are excluded from per-channel standardization and zero-filled before the recognition network.
Additionally, 14 trials flagged by a joint variance and template-correlation criterion as having residual onset artefacts on any kept channel beyond the active stimulation pair are dropped, leaving 423 trials.

\paragraph{Splits.}
Every trial is a single pulse at one site, so we stratify by stimulation site.
We hold out 10 trials for validation and 30 for testing, and train on the remaining 383.
The test set has two tiers: a held-out-site tier with all 10 trials stimulating the right-hippocampal pair RC3-RC4 (clinically labelled irritative zone), excluded from training and validation; and a held-in tier with the remaining 20 test trials from sites seen during training.

\subsection{Model Architecture}
\label{app:huang_arch}

The models follow the macaque setup (\Appref{app:kiani_arch}) with the modifications below; linear and nonlinear variants differ only in their dynamics parameterization.

\paragraph{Likelihood.}
Observations are continuous voltages, so we use a multivariate Gaussian likelihood $p(\vy_t \mid \vx_t) = \mathcal{N}(f_\theta(\vx_t), \mL\mL^\top)$ with a learned full covariance: $\mL$ is lower triangular with per-channel scales on the diagonal and learned off-diagonal entries below.

\begin{figure}[h]
  \centering
  \includegraphics[width=\linewidth]{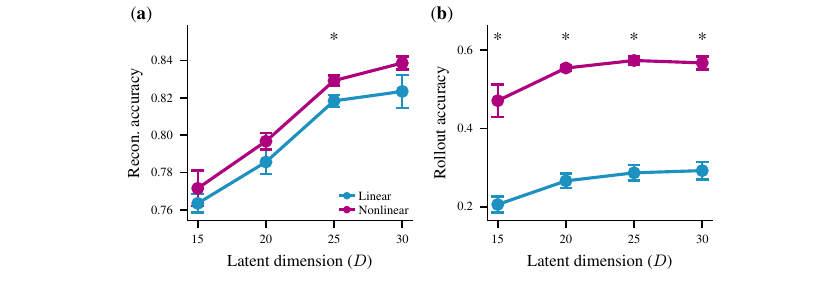}
  \caption{\textbf{Accuracy versus latent dimension $D$.}
  (a)~Reconstruction and (b)~rollout accuracy for linear (blue) and nonlinear (magenta) iSSMs; mean $\pm$ s.e.m.\ across 10 seeds.
  Asterisks mark significant differences ($p < 0.05$, FDR-corrected).}
  \label{fig:huang_d_sweep}
\end{figure}

\paragraph{Dynamics.}
As in \Appref{app:kiani_arch}, except for the $\mB$ parameterization (see below); we sweep $D \in \{15, 20, 25, 30\}$ and report $D{=}25$ in the main text.
The rollout advantage of the nonlinear iSSM holds at every $D$ (\Figref{fig:huang_d_sweep}).

\paragraph{Site-embedding intervention matrix.}
Unlike the macaque models, $\mB$ is not a free parameter: columns are predicted from electrode coordinates via a site-embedding network.
Let $\vs_m \in \R^3$ be the midpoint of bipolar site $m$ in the electrode coordinates provided with the dataset (\texttt{electrodes.tsv}).
We center and isotropically scale the coordinates (root-mean-square deviation over all sites and axes) and map them through a shared MLP:
\begin{equation}
    \vb_m = \mathrm{MLP}_\kappa\!\left(\frac{\vs_m - \bar{\vs}}{\sigma}\right) \in \R^D,
    \qquad
    \mB = \left[\, \vb_1, \ldots, \vb_M \,\right] \in \R^{D \times M}.
\end{equation}
$\mathrm{MLP}_\kappa$ has one hidden layer of 32 units with GeLU activation; the output layer is initialized with Gaussian weights scaled by $0.1$. This parameterization of $\mB$ by coordinates lets us predict responses at sites that were never stimulated during training, and, more speculatively, at sites that were never stimulated at all (\Appref{app:huang_virtual_stim}).
We note that this goes beyond our identifiability results, which constrain the input map only on the support of the observed inputs (\Remref{rem:input_map_support}). Generalisation to unseen sites instead relies on the smoothness of the coordinate parameterisation, which we evaluate empirically.

\paragraph{Emission.}
We use the residual-MLP emission of \Appref{app:toy_issm_arch} with hidden dimension $H_\mathrm{em}{=}128$, three residual blocks, and dropout probability $0.1$.

\paragraph{Recognition.}
As in \Appref{app:kiani_arch}, with $H_\mathrm{rec}{=}64$ and input dropout probability $0.2$.

\subsection{Training}
\label{app:huang_training}

Training follows \Appref{app:kiani_training}, with learning rate $3\times10^{-3}$, batch size 32, and no training-time $\mB$-sparsity warmup (since $\mB$ is computed from coordinates; the forward-pass threshold $\tau_B$ still applies).
We train 10 random seeds per latent dimension.
For the nonlinear model reported in the main text ($D=25$), the unordered-pairs estimate (\Appref{app:kiani_up}) is $\mathrm{up\_frac}=44.3 \pm 3.3\%$ (mean $\pm$ s.e.m. across 10 seeds).
Evaluation follows \Appref{app:real_eval}, with a fixed encoder window of 12 pre-stimulation timesteps ($\approx 47\,\mathrm{ms}$ at $\Delta t{=}3.9\,\mathrm{ms}$).

\subsection{Virtual Stimulation of Never-Stimulated Sites}
\label{app:huang_virtual_stim}

We synthesize 77 candidate bipolar pairs from adjacent recorded contacts on the same lead (33 never stimulated in the data), annotate each with the most severe clinical label of its two contacts, and group them into epileptogenic ($n{=}26$; seizure-onset, early-propagation, or irritative zone) and healthy ($n{=}51$) pairs.
Labels enter only at analysis time.
For each pair we obtain a virtual column $\vb$ from its midpoint coordinates and run two rollouts from each of 100 initial states (sampled from encoded latents at pulse onset across trials): one with a unit impulse along $\vb$ at the first step, one autonomous.
Then we difference the two to remove shared baseline drift.
Both trajectories are decoded and we take $\lVert \Delta \vy(t) \rVert$, averaged over initial states over a 50-step (195~ms) window.
From this curve we report the peak (response amplitude) and the mean from 100~ms to window end (response persistence).
We repeat this analysis for each of the ten random seeds from model training and aggregate the resulting readouts before testing.
Epileptogenic and healthy pairs are compared with Mann-Whitney $U$ tests, with Benjamini-Hochberg correction over the two readouts and two models.

\begin{figure}[h]
  \centering
  \includegraphics[width=\linewidth]{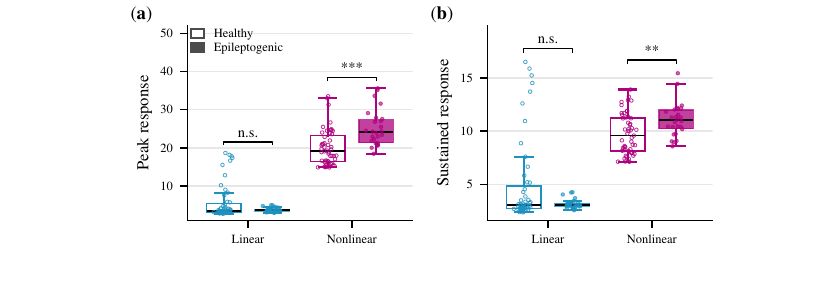}
  \caption{\textbf{Virtual stimulation of clinically annotated sites.}
  (a)~Peak and (b)~sustained network-wide response $\lVert \Delta \vy \rVert$ to virtual stimulation of healthy ($n = 51$) and epileptogenic ($n = 26$) bipolar pairs, under the linear and the nonlinear iSSM.
  Mann-Whitney $U$ tests, FDR-corrected over the two readouts and two models; $\ast\ast\ast$: $p < 0.001$.}
  \label{fig:huang_seizure}
\end{figure}

Under the nonlinear iSSM, epileptogenic pairs produce responses that are both larger (Cohen's $d = 1.08$, $p < 0.0001$) and more sustained ($d = 0.75$, $p = 0.0060$) than healthy pairs, matching the clinical expectation that seizure-onset tissue is hyperexcitable and hyperconnected; the linear iSSM shows neither effect (\Figref{fig:huang_seizure}).

Please note that we treat this as a pilot only. Input columns for many virtually stimulated sites are far from any sites actually stimulated in the dataset (\Figref{fig:huang}a), and outside the scope of our identifiability results (\Remref{rem:input_map_support}). Also, the present analysis uses only one subject and without independent replication.

\subsection{Computational Resources}
\label{app:huang_compute}

All experiments were run on CPU, with 4 cores and 6\,GB RAM per run for $\sim$2--4 hours.

\section{Stimulation Protocol Experiment Details}
\label{app:stim_details}

We provide full details for the stimulation protocol design experiments in \Secref{sec:stim}.
Both analyses use the bistable well of \Appref{app:dmm_systems} and vary one property of the intervention set at a time, holding the model class fixed.

\subsection{Diversity of Interventions}
\label{app:stim_diversity_details}

\paragraph{Data generation.}
Data generation follows \Appref{app:dmm_data}; only stimulator timing and amplitude parameters are varied, in a $2\times2$ factorial crossing pulse timing (regular or Poisson) with pulse amplitude (constant or diverse).
Under Poisson timing, pulses arrive as a Poisson process at rate $\lambda{=}0.75\,\mathrm{s}^{-1}$, each assigned to a uniformly random input channel.
Under regular timing, pulses cycle round-robin over the input channels with a fixed rest interval $d_\mathrm{rest} = 1/\lambda - d_\mathrm{stim} = 1.083\,\mathrm{s}$, so the two timings deliver the same expected number of pulses and differ only in their placement.
Under diverse amplitudes, each pulse amplitude is drawn independently from $\mathcal{U}[-3, 3]$.
Under constant amplitudes, a single amplitude is drawn once per run, with magnitude $\mathcal{U}[1.5, 3]$ and random sign, and used for every pulse in the run; amplitude and sign only vary across seeds.

\paragraph{Training.}
We train nonlinear iSSM only. Model architecture and training follow \Appsref{app:toy_issm_arch}{app:toy_issm_training}, with the settings of \Appref{app:dmm_training}. We train 10 random seeds for each of the four cells.

\paragraph{Computational resources.}
All runs were performed on CPU (2 cores, 2\,GB RAM).

\subsection{Replication on Point and Line Attractors}
\label{app:stim_point_line}

\begin{figure}[h]
    \centering
    \includegraphics[width=\linewidth]{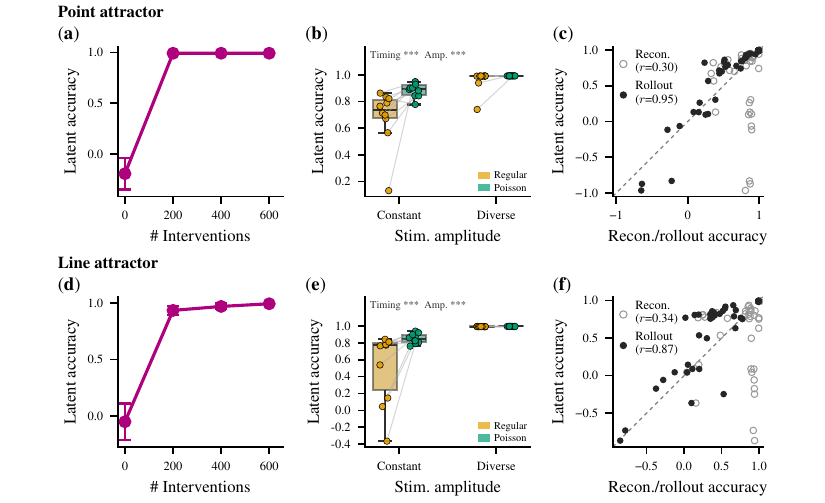}
    \caption{\textbf{Intervention-design trends replicate on point and line attractors.}
    Top row (a-c): point attractor; bottom row (d-f): line attractor.
    Panel layout matches \Figref{fig:interventions}: (a,d)~latent accuracy versus intervention count; (b,e)~amplitude-timing factorial (10 seeds); (c,f)~latent accuracy versus rollout (black) and reconstruction (gray); dashed line: identity.}
    \label{fig:interventions_point_line}
\end{figure}

We repeat the intervention-count sweep and the amplitude-timing factorial of \Secref{sec:stim} on the point and line attractor systems of \Appref{app:dmm_systems}, with the same training protocol as above.
The qualitative trends match the bistable well (\Figref{fig:interventions}): latent accuracy rises with more interventions, diverse amplitudes and irregular timing improve recovery, and rollout accuracy tracks latent accuracy more closely than reconstruction does (\Figref{fig:interventions_point_line}).

\section{Intervention Hardness}
\label{app:intervention_hardness}

\Thmref{thm:block_identifiability} assumes hard interventions that decouple driven latents from the past state (Eq.~\eqref{eq:general_issm_x}). We view intervention hardness as a gradual property that depends on the particular intervention technique. We therefore empirically test the robustness of latent identification to gradual violation of this assumption. 

We sweep the data-generating intervention from a purely additive drive ($\alpha=0$) to a perfectly hard clamp ($\alpha=1$) in the bistable well system. We compared the iSSM against a standard SSM and an input-conditioned variant whose nonlinear drift takes $\vu_t$ as an argument -- the most expressive of the three, as its hypothesis class contains every $\alpha$ regime.

\begin{figure}[h]
    \centering
    \includegraphics[width=\linewidth]{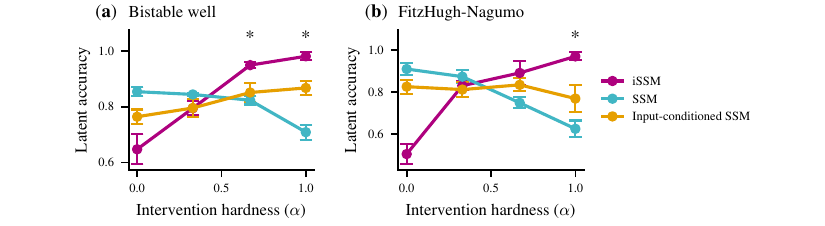}
    \caption{\textbf{Latent accuracy under softened interventions and model comparison.}
    Sweep over the intervention hardness parameter $\alpha$ of the data-generating process, from a purely additive drive ($\alpha=0$) to a hard clamp ($\alpha=1$), comparing nonlinear iSSM to two baselines.
    (a)~Bistable well (\Appref{app:hardness_bistable}).
    (b)~FitzHugh-Nagumo replication (\Appref{app:hardness_fhn}).
    Asterisks indicate $\alpha$-regimes where iSSM significantly outperforms both baselines ($p<0.05$).
    }
    \label{fig:baselines}
\end{figure}

On the bistable well (\Figref{fig:baselines}a), latent accuracy reveals two patterns.
First, iSSM identification improves gradually, not stepwise, as interventions harden; though it is poor under purely additive drives ($\alpha=0$) when the hardness assumption is completely violated.
Second, across all $\alpha$, only iSSM ever recovers the true latents near-perfectly: at $\alpha=1$ it reaches $0.98$, significantly above either baseline measured in its own ideal regime ($p=0.002$).
We replicate this analysis on FitzHugh-Nagumo dynamics (\Figref{fig:baselines}b), finding qualitatively similar results.

\subsection{Bistable Well}
\label{app:hardness_bistable}

\paragraph{Data generation.}
Data generation follows \Appref{app:dmm_data}, with 40 training and 10 validation trajectories (batch size 20), and with the hardness of the data-generating intervention swept.
Let $\vx_{t+1}^\mathrm{free} = g^*(\vx_t) + \vepsilon_t$ denote the state that the system would have reached without stimulation, where $g^*$ is the true one-step transition map.
At every driven latent entry we blend the clamp with these free dynamics,
\begin{equation}
    \vx_{t+1} = \vm_t \odot \vx_{t+1}^\mathrm{free}
    + (\vone - \vm_t) \odot \big( \mB \vu_t + (1-\alpha)\, \vx_{t+1}^\mathrm{free} \big),
    \label{eq:alpha_blend}
\end{equation}
with the intervention mask $m_{t,j} = \ind\{(\mB\vu_t)_j = 0\}$ of Eq.~\eqref{eq:general_issm_x}; undriven entries are unaffected.
At $\alpha{=}1$ the driven entry is overwritten by $\mB\vu_t$, recovering the hard intervention that Eq.~\eqref{eq:general_issm_x} assumes; at $\alpha{=}0$ the input is added on top of the free dynamics and the driven latent retains its dependence on the past state.
We sweep $\alpha \in \{1, 0.67, 0.33, 0\}$.
All other data properties are held at the values of \Appref{app:dmm_data}.

\paragraph{Model variants.}
The three models differ only in how the input $\vu_t$ enters the latent transition; emission, recognition, likelihood and initial-condition modules are identical across arms.
Writing $\mathrm{MLP}_\phi$ for the drift network of \Appref{app:toy_issm_arch}, the transitions are
\begin{align}
    \text{nonlinear iSSM:} && \vx_{t+1} &= \vm_t \odot \mathrm{MLP}_\phi(\vx_t) + \mB \vu_t + \vepsilon_t, \label{eq:baseline_m1}\\
    \text{standard SSM:} && \vx_{t+1} &= \mathrm{MLP}_\phi(\vx_t) + \mB \vu_t + \vepsilon_t, \label{eq:baseline_m2}\\
    \text{input-conditioned SSM:} && \vx_{t+1} &= \mathrm{MLP}_\phi([\vx_t, \vu_t]) + \vepsilon_t, \label{eq:baseline_m3}
\end{align}
with $m_{t,j} = \ind\{(\mB\vu_t)_j = 0\}$ as above.

Nonlinear iSSM is misspecified at $\alpha\neq 1$; standard SSM is misspecified at $\alpha \neq 0$; the input-conditioned SSM is never misspecified.

\paragraph{Training.}
Model architecture and training follow \Appsref{app:toy_issm_arch}{app:toy_issm_training}, with the settings of \Appref{app:dmm_training}. We train 10 random seeds for each (model, $\alpha$) condition.

\paragraph{Evaluation.}
Evaluation follows \Appref{app:toy_eval}, with one exception: since the input-conditioned SSM has no~$\mB$, we index-align latents from latent trajectories rather than $\mB$ columns for all models in this experiment.

\paragraph{Computational resources.}
All runs were performed on CPU (4 cores, 2\,GB RAM for training and 4\,GB for evaluation), with per-seed training cost comparable to \Appref{app:dmm_compute}.

\subsection{FitzHugh-Nagumo Replication}
\label{app:hardness_fhn}

\paragraph{Data generation.}
Data generation follows \Appref{app:fhn_data}, with the intervention hardness blended as in Eq.~\eqref{eq:alpha_blend} and swept over $\alpha \in \{1, 0.67, 0.33, 0\}$.
All other data properties match \Appref{app:fhn_data}.

\paragraph{Model variants, training, and evaluation.}
We use the same three model classes (Eqs.~\eqref{eq:baseline_m1}--\eqref{eq:baseline_m3}), training protocol (\Appref{app:toy_issm_training}), and evaluation protocol (\Appref{app:toy_eval}, with index alignment for the input-conditioned SSM) as in \Appref{app:hardness_bistable}.
We train 10 random seeds for each (model, $\alpha$) condition.

\paragraph{Computational resources.}
All runs were performed on CPU (4 cores, 2\,GB RAM for training and 4\,GB for evaluation), with per-seed training cost comparable to \Appref{app:fhn_compute}.

\end{document}